\documentclass[sigconf]{acmart}

\AtBeginDocument{%
  \providecommand\BibTeX{{%
    \normalfont B\kern-0.5em{\scshape i\kern-0.25em b}\kern-0.8em\TeX}}}

\acmConference[Conference acronym 'XX]{Make sure to enter the correct
  conference title from your rights confirmation emai}{June 03--05,
  2018}{Woodstock, NY}
\acmISBN{978-1-4503-XXXX-X/18/06}

\usepackage{float}
\usepackage{newfloat}
\usepackage{multirow}
\usepackage{makecell}
\usepackage{xcolor}         
\usepackage{colortbl}
\usepackage{dsfont}
\usepackage{amsmath}
\usepackage{enumitem}
\usepackage{algpseudocode}
\usepackage[ruled,lined]{algorithm2e}
\usepackage{tcolorbox} 
\usepackage[normalem]{ulem}
\useunder{\uline}{\ul}{}

\usepackage{wrapfig}
\usepackage{subcaption}

\usepackage[normalem]{ulem}
\useunder{\uline}{\ul}{}

\definecolor{myyellow}{rgb}{1,1, 0.6}
\definecolor{myorange}{rgb}{1, 0.8, 0.6}
\definecolor{myred}{rgb}{1, 0.6, 0.6}
\definecolor{second}{HTML}{FFDAB9}
\definecolor{best}{HTML}{FFC1C1}
\usepackage{multirow}
\usepackage{colortbl}
\usepackage[normalem]{ulem}
\usepackage{bbding}
\useunder{\uline}{\ul}{}
\usepackage{stackrel}
\usepackage{extarrows}
\usepackage{todonotes}
\copyrightyear{2025}
\acmYear{2025}
\setcopyright{acmlicensed}\acmConference[MM '25]{Proceedings of the 33rd ACM International Conference on Multimedia}{October 27--31, 2025}{Dublin, Ireland}
\acmBooktitle{Proceedings of the 33rd ACM International Conference on Multimedia (MM '25), October 27--31, 2025, Dublin, Ireland}
\acmDOI{10.1145/3746027.3754895}
\acmISBN{979-8-4007-2035-2/2025/10}

\renewcommand{\todo}[1]{\iffalse #1 \fi{\color{blue} \textbf{[TODO]}}}

\newcommand{\leftrarrows}{\mathrel{\raise.9ex\hbox{\oalign{%
  $\scriptstyle\leftarrow$\cr
  \vrule width0pt height.5ex$\hfil\scriptstyle\relbar$\cr}}}}
\newcommand{\lrightarrows}{\mathrel{\raise.9ex\hbox{\oalign{%
  $\scriptstyle\relbar$\hfil\cr
  $\scriptstyle\vrule width0pt height.5ex\smash\rightarrow$\cr}}}}
\newcommand{\Rrelbar}{\mathrel{\raise.9ex\hbox{\oalign{%
  $\scriptstyle\relbar$\cr
  \vrule width0pt height.5ex$\scriptstyle\relbar$}}}}

\newcommand{\longlonglongrightarrows}{\joinrel\relbar\joinrel\relbar\joinrel\relbar\joinrel\relbar\joinrel\relbar\joinrel\relbar\joinrel\relbar\joinrel\relbar\joinrel\relbar\joinrel\relbar\joinrel\longrightarrow}

\newcommand{\longlongleftrightarrows}{\leftrarrows\joinrel\Rrelbar\joinrel\Rrelbar\joinrel\Rrelbar\joinrel\Rrelbar\joinrel\Rrelbar\joinrel\Rrelbar\joinrel\Rrelbar\joinrel\Rrelbar\joinrel\Rrelbar\joinrel\lrightarrows}

\newcommand{\longlonglonglongleftrightarrows}{\leftrarrows\joinrel\Rrelbar\joinrel\Rrelbar\joinrel\Rrelbar\joinrel\Rrelbar\joinrel\Rrelbar\joinrel\Rrelbar\joinrel\Rrelbar\joinrel\Rrelbar\joinrel\Rrelbar\joinrel\Rrelbar\joinrel\Rrelbar\joinrel\Rrelbar\joinrel\Rrelbar\joinrel\Rrelbar\joinrel\Rrelbar\joinrel\Rrelbar\joinrel\Rrelbar\joinrel\Rrelbar\joinrel\Rrelbar\joinrel\Rrelbar\joinrel\Rrelbar\joinrel\Rrelbar\joinrel\lrightarrows}

\newcommand{\problemFull}{real-time distribution shift on devices}
\newcommand{\problem}{on-device real-time data distribution shift}
\newcommand{\method}{Persona}
\newcommand{\moduleA}{Prototype Model}
\newcommand{\moduleAbrief}{PM}
\newcommand{\moduleB}{Parameter Editor}
\newcommand{\moduleBbrief}{PE}
\newcommand{\modelGradient}{Parameter Editing Matrix}
\newcommand{\modelDevice}{DAM}
\newcommand{\methodsName}{Personalized}

\newtheorem{assumption}{Assumption}

\newtheorem{lemma}{Lemma}

\begin{document}

\title{Tackling Device Data Distribution Real-time Shift via Prototype-based Parameter Editing}
\author{Zheqi Lv}
\affiliation{%
  \institution{Zhejiang University}
  \city{Hangzhou}
  \country{China}}
\authornote{Contributed equally.}
\email{zheqilv@zju.edu.cn}

\author{Wenqiao Zhang}
\affiliation{%
  \institution{Zhejiang University}
  \city{Hangzhou}
  \country{China}}
\authornotemark[1]
\email{wenqiaozhang@zju.edu.cn}

\author{Kairui Fu}
\affiliation{%
  \institution{Zhejiang University}
  \city{Hangzhou}
  \country{China}}
\email{fukairui.fkr@zju.edu.cn}

\author{Qi Tian}
\affiliation{%
  \institution{Zhejiang University}
  \city{Hangzhou}
  \country{China}}
\email{tianqics@zju.edu.cn}

\author{Shengyu Zhang}
\affiliation{%
  \institution{Zhejiang University}
  \city{Hangzhou}
  \country{China}}
\authornote{Corresponding authors.}
\email{sy_zhang@zju.edu.cn}

\author{Jiajie Su}
\affiliation{%
  \institution{Zhejiang University}
  \city{Hangzhou}
  \country{China}}
\email{sujiajie@zju.edu.cn}

\author{Jingyuan Chen}
\affiliation{%
  \institution{Zhejiang University}
  \city{Hangzhou}
  \country{China}}
\email{jingyuanchen@zju.edu.cn}

\author{Kun Kuang}
\affiliation{%
  \institution{Zhejiang University}
  \city{Hangzhou}
  \country{China}}
\authornotemark[2]
\email{kunkuang@zju.edu.cn}

\author{Fei Wu}
\affiliation{%
  \institution{Zhejiang University}
  \city{Hangzhou}
  \country{China}}
\email{wufei@zju.edu.cn}

\renewcommand{\shortauthors}{Zheqi Lv et al.}



\begin{abstract}
\label{sec:abstract}

The on-device real-time data distribution shift on devices challenges the generalization of lightweight on-device models. This critical issue is often overlooked in current research, which predominantly relies on data-intensive and computationally expensive fine-tuning approaches. To tackle this, we introduce Persona, a novel personalized method using a prototype-based, backpropagation-free parameter editing framework to enhance model generalization without post-deployment retraining. Persona employs a neural adapter in the cloud to generate a parameter editing matrix based on real-time device data. This matrix adeptly adapts on-device models to the prevailing data distributions, efficiently clustering them into prototype models. The prototypes are dynamically refined via the parameter editing matrix, facilitating efficient evolution. Furthermore, the integration of cross-layer knowledge transfer ensures consistent and context-aware multi-layer parameter changes and prototype assignment. Extensive experiments on vision task and recommendation task on multiple datasets confirm Persona's effectiveness and generality.
\end{abstract}

\maketitle

\section{Introduction}
\label{sec:introduction}

\begin{figure*}[t]
  \centering
\includegraphics[width=0.95\linewidth]{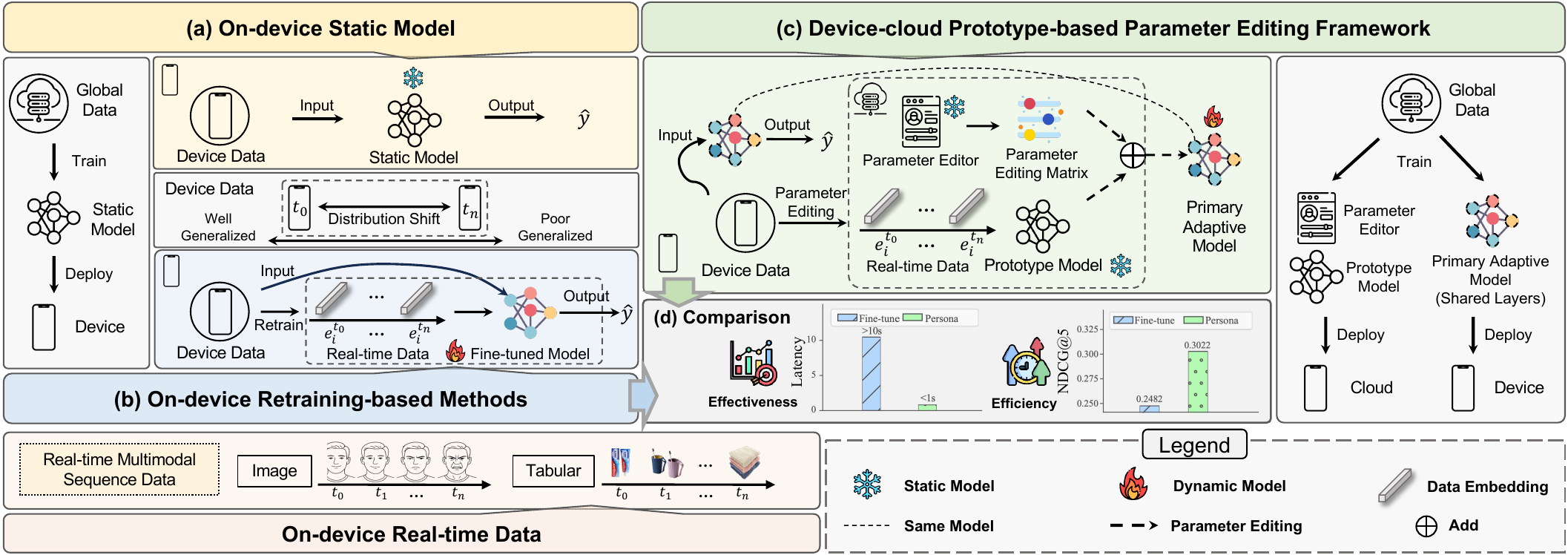}
\vspace{-0.3cm}
  \caption{ (a) describes the on-device static model. (b) describes traditional paradigm which can be used to solve the \problem{}. (c) illustrates our \method{}. In (a), (b), and (c), the term "Device Data" refers to real-time sequential data from multiple modalities collected on the device across various scenarios, such as image modality sequences (e.g., facial expression recognition) and tabular modality sequences (e.g., user behavior modeling). (d) is the comparison of baselines and \method{} (Time Consumption:
1s (\method{}) $\ll$ 10s (Fine-tuning)), NDCG@5: 0.3022 (\method{}) $>$ 0.2482 (Fine-tuning)).
}
  \label{fig:introduction}
\vspace{-0.35cm}
\end{figure*}

Deep neural networks often learn from vast global data collected from devices to create cloud-based models~\cite{ref:vggnet,ref:resnet,ref:liu2019roberta,ref:ensemble_mazari2024bert,ref:deepfm}. 
This cloud-centric approach could introduce latencies between on-device data/request generation and the delivery of prediction results, leading to missed opportunities for device participation~\cite{DBLP:conf/kdd/YangHXLYL22}. 
To mitigate this issue, it is common to deploy static, cloud-pretrained models directly on devices, as illustrated in Figure~\ref{fig:introduction}(a)\cite{ref:mobilenetv3,ref:sasrec}. Nevertheless, these static models often struggle to adapt to dynamically changing local environments, such as altering perspectives in autonomous vehicles or evolving user preferences in recommender systems. This inflexibility potentially undermines the efficacy of real-time decision-making and degrades user experiences\cite{DBLP:conf/kdd/YangHXLYL22}~\cite{ref:multi_modal_personalized}. Consequently, there is an increasing necessity for investigating real-time generalization, \textit{i.e.}, models which can dynamically generalize to reflect ongoing changes and address \problem{}.

A straightforward solution is to drive on-device generalization involves instant fine-tuning, as illustrated in Figure~\ref{fig:introduction}(b). Recognizing the challenge of sparse labeled data on many devices, which might precipitate overfitting when directly applying fine-tuning, recent studies have explored methods to synthesize or extract distribution-specific data samples from heterogenous sources, such as cloud storage or other devices~\cite{ref:device_cloud2,ref:devicefinetune,ref:devicefinetune2,ref:devicefinetune3,ref:dccl}. Despite these advancements, a significant challenge remains: these methods typically demand extensive computational resources, and could hardly meet the requirements of \textit{real-time} generation, \textit{i.e.}, to swiftly adjust to new distributions and deliver prompt responses.

In this paper, our goal is to continuously provide real-time, generalized  deep learning models, balancing adaptability and computational efficiency. To this end, as shown in the Figure~\ref{fig:introduction}(c), we propose a Prototype-based Parameter Editing Framework (\emph{abbr.} \textbf{\method{}}), to address the aforementioned \problem{}. At its core, \method{} incorporates a \moduleB{}, which ingeniously maps real-time device data into \modelGradient{}. The matrix transforms a shared, unified deep learning model into a adaptive model to better serve for the current on-device data distribution. Remarkably, this process necessitates once inference of \moduleB{}s, fueling efficient and responsive generalization.

Despite the global optimality of the universally learned deep learning model, it may not uniformly cater to the disparate needs of individual devices. Our empirical study, as elucidated in Section \ref{subsubsec:effectiveness_evaluation}, lends credence to our proposition: the \moduleB{}'s efficacy is proportional to the proximity of generalization needs to the shared deep learning model. 
In this regard, \method{} adopts the prototype-editor paradigm. The initial \modelGradient{} facilitate the clustering of devices and the construction of multiple shared \moduleA{}s, each corresponding to a uniquely crafted editor. These editors perform the twofold task of generalizing device models through the application of the generated \modelGradient{}, and enhancing the dynamic assignment of device models to \moduleA{}s. The efficacy of our framework hinges on the delicate equilibrium maintained between these dual roles of generalization and assignment. The proposed dynamic device assignment strategy assigns devices with the \moduleA{} necessitating the least adaptation. 
The core principle behind our strategy is straightforward yet effective: in situations where the on-device distribution undergo a substantial transformation from its prior state, we facilitate adaptation from another \moduleA{} that is optimally equipped to understand new device data. Furthermore, we promote knowledge transfer across different layers, a mechanism that assures consistency in multi-layer \modelGradient{}. This consistency is paramount to maintaining stability and performance in prototype assignment, as inconsistent biases across layers could result in conflicting device-prototype assignment.

In summary, this work makes the following three key contributions: (1) Our investigation is among the early efforts to tackle \problemFull{} without the necessity for substantial on-device computational resources and labeled data. (2) We propose and implement the Prototype-based Parameter Editing framework, a novel approach that leverages the power of neural editors to transform \moduleA{}s for generalization without any backward propagation. We further introduce the dynamic and consistent prototype assignment learning strategies, contributing to better adaptability and effectiveness. (3) We conducted extensive experimental studies on tasks of multiple modality including  multiple datasets to verify the  superiority of \method{}. The experiments demonstrate that the \method{} achieves significant improvements, surpassing existing SOTA approaches. Moreover, \method{} exhibits generalizability across diverse datasets, confirming its potential to address the challenges associated with device data distribution.

\section{Related Works}
\noindent\textbf{Device-Cloud Collaboration.}
Conventional cloud-based learning has been widely applied~\cite{zhang2024modality, zhang2025personalized, liu2023category_prototype, wangneural, wang2023deconfounded_mm, li2023your_mm, su2023personalized, su2025distilling_mmrec, wu2025embracing,zhu2025graphclip,su2023enhancing,zhu2025graphclip,zhang2021mining_mmrec,khani2023recl}, but with the rapid increase in the number of smart devices and their growing computational resources, Device-cloud collaboration is gaining more and more attention, blending the advantages of cloud-based and on-device machine learning~\cite{ref:device_cloud2,ref:devicefinetune2,ref:devicefinetune3,ref:devicefinetune4,ref:device_cloud_cv,ref:device-cloud_yolo,ref:device_cloud_walle,ref:device-cloud_adaptation,ref:devicefinetune,ref:device_cloud,ref:zhang_device_cloud,lv2025collaboration,fu2025forward,ji2025backpropagation,fu2024diet}.
DCCL~\cite{ref:dccl} improves the generalization ability of the device model by add an adaptive layer into the device model and update continuously, MPDA~\cite{ref:device_cloud2} improves performance by training on cloud-extracted samples similar to those on the device. In addition, \cite{ref:devicefinetune} and \cite{ref:devicefinetune4} made preliminary attempts at low-resource scenario fine-tuning in some contexts. Additionally, \cite{ref:device_cloud,ref:lv2024intelligent} focus on whether there is a distribution changing on device.

\noindent\textbf{Dynamic Neural Network.}
The Dynamic neural network focus on generating dynamic linear layers~\cite{ref:hypernetwork_continual_learning,ref:hypernetwork_graph,ref:hypernetwork_meta_learning,ref:hypernetwork_federated_learning,ref:hypernetwork_hyperstyle,ref:hypernetwork_hyperinverter,tang2024modelgpt,lv2024semantic} and dynamic CNN~\cite{ref:dfn_dynamic_conv,ref:dfn_condconv,ref:dfn_instance_segmentation,ref:dfn_lightweight,ref:dfn_semantic_segmentation,ref:dfn_unknown1,ref:dfn_unknown2,ref:dfn} according to the samples, unlike a traditional static neural network for all samples. The main goal of these methods is to overcome the negative impact of distributional shifts on model generalization~\cite{HTCL_ood,Li_2025_CVPR_ood,yang2024explain_seqood,liu2024rethinking_ood}.
\cite{ref:dfn_lightweight} splits dynamic neural network into channel and point convolutions for lightweight application in more network layers. Both CondConv \cite{ref:dfn_condconv} and DynamicConv \cite{ref:dfn_dynamic_conv} use attention to aggregate filters. Dynamic neural network is also used in semantic segmentation (DMNet \cite{ref:dfn_semantic_segmentation}) and instance segmentation (SoloV2 \cite{ref:dfn_instance_segmentation}).

\noindent\textbf{On-device Deep Learning Models.}
Deep learning models excel in multi-modal tasks. In visual tasks, ResNet~\cite{ref:he2016resnet} addresses the vanishing gradient problem, while SqueezeNet~\cite{ref:squeezenet}, ShuffleNet~\cite{ref:shufflenet,ref:shufflenetv2}, and MobileNet~\cite{ref:mobilenet,ref:mobilenetv2,ref:mobilenetv3} enhance efficiency~\cite{ref:efficientnet,ref:ghostnet}. In recommendation tasks, DeepFM~\cite{ref:deepfm} and LightGCN~\cite{ref:he2020lightgcn} improve collaborative filtering by capturing nonlinear features. 
Models like DIN~\cite{ref:din}, GRU4Rec~\cite{ref:gru4rec}, and SASRec~\cite{ref:sasrec} can also achieve real-time computation on devices by dispatching item embeddings in the candidate list from the cloud.

\section{Methodology}
\label{sec:method}

Here we introduce \method{}. \textbf{Part of the \textit{methodology} and \textit{theoretical proof} can be found in the Appendix.}
\subsection{Problem Formulation and Notations}
\begin{sloppypar}
In the device-cloud system, it is necessary to collect historical data $\mathcal{D}_H=\{\mathcal{X}^{(i)}, \mathcal{Y}^{(i)}\}_{i=1}^{\mathcal{N}_H}$ for a period of time from different devices $\{d^{(i)}\}_{i=1}^{\mathcal{N}_d}$, where $\mathcal{X}^{(i)}$ can represent data from multiple modalities (e.g., \emph{ID}, \emph{images}, etc.). The models $\mathcal{M}^{(0)}_\mathcal{G}, \{\mathcal{M}^{(j)}_\mathcal{G}\}_{j=1}^{\mathcal{N}_\mathcal{M}}, \mathcal{M}^{(0)}_\mathcal{F}, \{\mathcal{M}^{(j)}_\mathcal{F}\}_{j=1}^{\mathcal{N}_\mathcal{M}}$, and $\mathcal{M}_\mathcal{F}$ are obtained by training on the cloud based on $\mathcal{D}_H$, the parameters of them is represented by $\Theta^{(0)}_\mathcal{G}, \{\Theta^{(j)}_\mathcal{G}\}_{j=1}^{\mathcal{N}_\mathcal{M}}, \Theta^{(0)}_\mathcal{F}, \{\Theta^{(j)}_\mathcal{F}\}_{j=1}^{\mathcal{N}_\mathcal{M}}, \Theta_\mathcal{F}$. We define $\Theta_\mathcal{F}=\{\Theta_b, \Theta_c\}, \Theta^{(0)}_\mathcal{F}=\{\Theta^{(0)}_b, \Theta^{(0)}_c\}$, where $\Theta_b$ and $\Theta_c$ represent the parameters of the $\mathcal{M}_\mathcal{F}$'s shared layers and adaptive layers respectively. 
The parameters of the adaptive layers $\Theta_c^{(0)}$ is generated by $\mathcal{M}^{(0)}_\mathcal{G}$. $\mathcal{M}_\mathcal{F}$ is deployed on the device and inferences based on real-time data $\mathcal{D}_R=\{\mathcal{X}^{(i)}\}_{i=1}^{\mathcal{N}_R}$ as device models usually do.
What needs to be emphasized is that $\mathcal{M}_\mathcal{F}$ is a model deployed and used for inference on the device, and its $\Theta_b$ is fixed, and $\Theta_c$ changes in real time with the data distribution. $\mathcal{M}_\mathcal{F}^{(0)}$ is a model with fixed parameters, $\mathcal{M}_\mathcal{G}^{(0)}$ generates \modelGradient{} based on the parameters of $\mathcal{M}_\mathcal{F}^{(0)}$ and aggregates them with the parameters of $\mathcal{M}_\mathcal{F}^{(0)}$ to get the adaptive layer parameter $\Theta_c$ of $\mathcal{M}_\mathcal{F}$ and send it to the device, and then $\mathcal{M}_\mathcal{F}$ updates $\Theta_c$.
However, due to the limitations of $\{\mathcal{M}_\mathcal{G}^{(0)}, \mathcal{M}_\mathcal{F}^{(0)}\}$, we designed $\{\mathcal{M}^{(j)}_\mathcal{G}, \mathcal{M}^{(j)}_\mathcal{F}\}_{j=1}^{\mathcal{N}_\mathcal{M}}$ to dynamically generate the parameters based on real-time data $\mathcal{D}_R$. $\mathcal{N}_d, \mathcal{N}_H, \mathcal{N}_R, \mathcal{N}_\mathcal{M}$ respectively denote the amount of devices, historical data, real-time data and \moduleAbrief{}. 
\end{sloppypar}

The update of device model $\mathcal{M}_\mathcal{F}$ can be regarded as several steps. 
(1) The device uploads real-time data $\mathcal{X}_{R^{(i)}}$ to the cloud. 
(2) $\{\mathcal{M}^{(j)}_\mathcal{G}\}_{j=1}^{\mathcal{N}_\mathcal{M}} $  generates \modelGradient{} $\{G^{(j)}\}_{j=1}^{\mathcal{N}_\mathcal{M}}$ based on multiple \moduleAbrief{}s $\{\mathcal{M}^{(j)}_\mathcal{F}\}_{j=1}^{\mathcal{N}_\mathcal{M}}$. 
(3) Assign models dynamically according to $\{G^{(j)}\}_{j=1}^{\mathcal{N}_\mathcal{M}}$, and select the group \moduleAbrief{} from $\{\mathcal{M}^{(j)}_\mathcal{F}\}_{j=1}^{\mathcal{N}_\mathcal{M}}$ and the group \moduleB{} (\moduleBbrief{}) from $\{\mathcal{M}^{(j)}_\mathcal{G}\}_{j=1}^{\mathcal{N}_\mathcal{M}} , $. 
(4) Suppose $j'$ is selected, then we aggregate $G^{(j)}$ generated by $\mathcal{M}^{(j')}_\mathcal{G}$ and $\mathcal{M}^{(j)}_\mathcal{F}$ into $\Theta_{\mathcal{F}}$.
(5) The device model $\mathcal{M}_\mathcal{F}$ update the parameters to $\Theta_{\mathcal{F}}$.
First, we construct a novel framework  \method{} to tackle the \problem{}.
\textbf{Single-Prototype \method{}} can be formulated as follows,

\noindent \textit{Device-Cloud Communication}:\\
\begin{equation}
\resizebox{0.47\textwidth}{!}{
$
\underbrace{
\mathcal{M}^{(0)}_{\mathcal{G}}
(\mathcal{X}_{R^{(i)}};\Theta^{(0)}_{\mathcal{G}})\bigoplus\mathcal{M}^{(0)}_{\mathcal{F}}}_{\rm{Global\ Cloud\ Model}}  
\stackrel[\text{Parameters }\Theta_c]{\text{Real-time Data }\mathcal{X}_{R^{(i)}}}{\longlongleftrightarrows}
\underbrace{
\mathcal{M}_{\mathcal{F}}
(\mathcal{X}_{R^{(i)}};\Theta_{\mathcal{F}}=\{\Theta_b, \Theta_c\})}_{\rm{Local\ Device\ Model}}.
 \label{eq:problem_formulation_collaboration_1}
$
}
\end{equation}

Furthermore, we propose \textbf{Multi-Prototype \method{}}, which exhibits a more effective \method{} scheme:

\noindent \textit{On-Cloud}:\\
\begin{equation}
\resizebox{0.47\textwidth}{!}{
$
\underbrace{
{\rm{Assign}}(\{\mathcal{M}_\mathcal{G}^{(j)}, \mathcal{M}_\mathcal{F}^{(j)}\}_{j=1}^{\mathcal{N}_\mathcal{M}}, \{G^{(i)}\}_{i=1}^{\mathcal{N}_\mathcal{M}})}_{\rm{Global\ Cloud\ Model}}
\stackrel[
\text{Group Model }
\{\mathcal{M}^{(j^{'})}_{\mathcal{G}}, \mathcal{M}^{(j^{'})}_{\mathcal{F}}\}]{\text{\modelGradient{} }\{G^{(i)}\}^{N_\mathcal{M}}_{i=1}}{\longlonglonglongleftrightarrows}
\underbrace{
\mathcal{M}^{(0)}_{\mathcal{G}}
(\mathcal{X}_{R^{(i)}};\Theta^{(0)}_{\mathcal{G}})}_{\rm{Global\ Cloud\ Model}}.
$
}
\end{equation}

\noindent \textit{Device-Cloud Communication}:\\
\begin{equation}
\resizebox{0.47\textwidth}{!}{
$
\underbrace{
\mathcal{M}^{(j^{'})}_{\mathcal{G}}
(\mathcal{X}_{R^{(i)}};\Theta^{(j^{'})}_{\mathcal{G}})\bigoplus\mathcal{M}^{(j^{'})}_{\mathcal{F}}}_{\rm{Global\ Cloud\ Model}}  
\stackrel[\text{Parameters }\Theta_c]{\text{Real-time Data }\mathcal{X}_{R^{(i)}}}{\longlongleftrightarrows}
\underbrace{
\mathcal{M}_{\mathcal{F}}
(\mathcal{X}_{R^{(i)}};\Theta_{\mathcal{F}}=\{\Theta_b, \Theta_c\})}_{\rm{Local\ Device\ Model}}.
 \label{eq:problem_formulation_collaboration_2_2}
$
}
\end{equation}

\begin{figure*}[t]
  \centering
\includegraphics[width=0.96\linewidth]{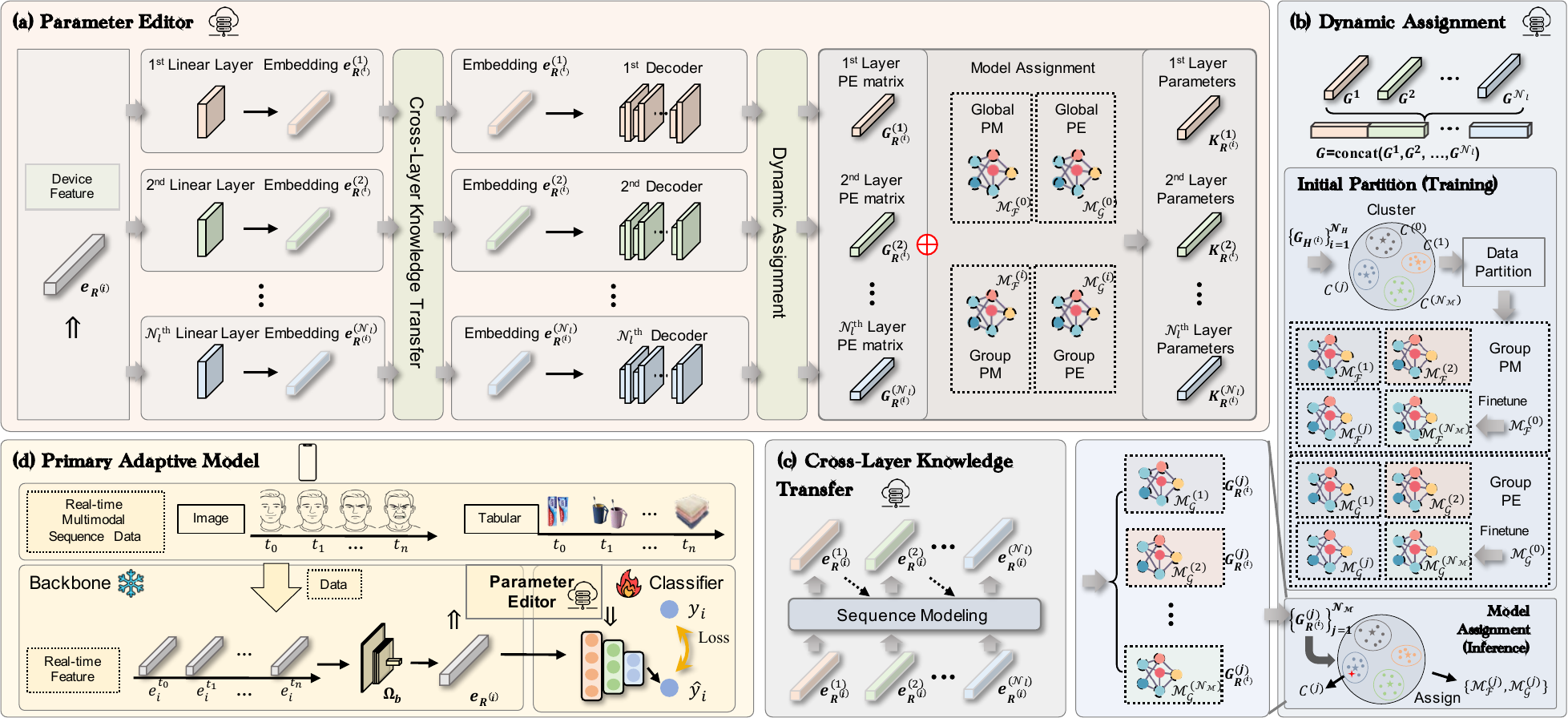}
\vspace{-0.25cm}
  \caption{
  \method{} Overview. (a) \moduleA{} includes shared layers and adaptive layers. The parameters of the adaptive layers are dynamically generated by the global \moduleBbrief{} $\mathcal{M}^{(0)}_{G}$ based on real-time data, producing \modelGradient{} to edit the global \moduleAbrief{} $\mathcal{M}^{(0)}_{F}$. (b) The partitioning and dynamic allocation algorithm allows the parameter editing unit to be finely grouped into global \moduleAbrief{} and global \moduleBbrief{}, resulting in group \moduleAbrief{} and group \moduleBbrief{}. (c) Cross-layer knowledge transfer pulls the \modelGradient{} of different layers for the same sample into one space, ensuring unified and stable partitioning of the samples.
  }
  \label{fig:architecture}
\vspace{-0.3cm}
\end{figure*}

\subsection{Single-Prototype \method{}}
In this section, we present \method{}. Figure~\ref{fig:architecture} illustrates the overview of \method{}.
\subsubsection{Primary Adaptive Model}
\label{subsec:primary_network}
In \method{}, we decouple the \textbf{On-\underline{D}evice \underline{A}daptive \underline{M}odel}~(\modelDevice{})~(Figure~\ref{fig:architecture}(d)) into ``Shared Layers'' and ``Adaptive Layers''. 
(1) Shared layers $\Omega$, whose parameters $\Theta_b$ learned from global data, use shared knowledge to generate representations for devices, with their parameters fixed after training, serving all devices. 
(2) The weights of the adaptive layer ${\Theta}_c$ are generated by \moduleBbrief{} based on real-time device data. That is, with the change in real-time device data, the parameters of the adaptive layer will change rapidly and in real-time during the inference stage.

The training process of \modelDevice{} is synchronized with the adaptive Weights Generator, and we will introduce its training procedure in Sec.~\ref{subsec:adaptive_weights_generator}.
After we obtained a well-trained \modelDevice{} with a backbone $\Omega_b$ and \moduleBbrief{} $\mathcal{M}_\mathcal{G}$, the inference procedure of the on-device \modelDevice{} can be computed as follows:
\begin{equation}
\begin{aligned}
\mathcal{L}_{H}^{(j)}=\sum_{i=1}^{\mathcal{N}_d} \sum_{j=1}^{ \mathcal{N}_{R^{(i)}}} 
l_{ce} (y^{(j)}_{H^{(i)}}, \Omega(x^{(j)}_{H^{(i)}};\Theta_b); \Theta_c:=\mathcal{G}(x^{(j)}_{H^{(i)}};\Theta_{\mathcal{G}})),
\end{aligned}
\label{eq:pam_pred}
\end{equation}

where $l_{ce}(\cdot)$ denotes the cross-entropy loss function which is used to fit the \modelDevice{} to the ground-truth. $\Omega(x^{(j)}_{H^{(i)}};\Theta_b)$ is the backbone extracting features from sample $x^{(j)}_{H^{(i)}}$.
$\Theta_b$ and $\Theta_c$ are the learnable parameters for the backbone and classifier, respectively.

\subsubsection{\moduleB{}}
\label{subsec:adaptive_weights_generator}
In this section, we introduce the \textbf{\moduleB{}}~(\moduleBbrief{}). Generally speaking, the bootstrapping philosophy of \moduleBbrief{} is to dynamically generate \modelGradient{} $g_i$ conditioned on the basic device model by different condition $z_i$, i.e., $W_i = \mathcal{G}(z_i)$ where $\mathcal{G}$ refers to the adaptive parameter generation network. Then the generated parameters are applied to the deep learning models, i.e., $y_i = \mathcal{F}(x_i;\mathcal{G}(z_i))$. Next, we present (1) how to design the condition $z_i$ for the different instance $i$ and (2) how to implement $\mathcal{G}$.

To generate the parameters for $n^{th}$ layer of ``adaptive Layers'' in the primary model, we develop a layer encoder to represent the $n^{th}$ layer parameters as an embedding $\boldsymbol{e}^{n}_{R^{(i)}}$. To model relationships of different layers, instead of constructing the one-to-one encoder-layer correspondence, the $\boldsymbol{e}^{n}_{R^{(i)}}$ share one encoder neck but use different linear layers to change the real-time data features.
\begin{align}
\label{eq:lightweight_encoder}
\vspace{-0.2cm}
    \boldsymbol{e}^{n}_{R^{(i)}} = {L}^{n}(\boldsymbol{e}_{R^{(i)}}), \quad \boldsymbol{e}_{R^{(i)}}={E}(\mathcal{X}_{R^{(i)}}), \forall n\in  \{1, 2, \cdots, \mathcal{N}_l\},
\vspace{-0.2cm}
\end{align}
where ${E}(\cdot)$ represents a general encoder. ${L}^{n}(\cdot)$ is a linear layer used to adjust $\boldsymbol{e}_{R^{(i)}}$ to the $n^{th}$ layer-specific feature. 

We treat the parameters and the \modelGradient{} as a matrix $K^{n}\in\mathbb{R}^{N^{n}_{\rm{in}}\times N^{n}_{\rm{out}}}$, where $N^{n}_{\rm{in}}$ and $N^{n}_{\rm{out}}$ represent the number of input neurons and output neurons of the $n^{th}$ layer of the on-device \modelDevice{}, respectively. 
The \moduleBbrief{} $g(\cdot)$ is used to generate \modelGradient{} based on the features extracted from real-time data for the \modelDevice{} by $G^{n}_{R^{(i)}} = g^{n}(\boldsymbol{e}^{n}_{R^{(i)}})$.
\begin{align}
\label{eq:kernal_generation_detail}
\begin{aligned}
    K^{n}_{R^{(i)}} &= {\rm{Clip}}(\Theta^{n}_\triangle)+ \Theta^{n} \\
    &= {\rm{Clip}}({\rm{Reshape}}(G^{n}_{R^{(i)}}; N^{n}_{\rm{in}}, N^{n}_{\rm{out}}); \mathcal{T}) + \Theta^{n},
\end{aligned}
\vspace{-0.2cm}
\end{align}
where $\mathcal{T}$ is a hyperparameter to control the threshold of the generated \modelGradient{}.
Note that since the parameters of adaptive layers for the pretrained model with fixed parameter $\Theta^{n}$ can map data points to the not-bad location in the feature space, we believe the optimal feature should be around this location. 

The training procedure of the proposed \method{} framework can thus be formulated as the following optimization problem:
\begin{equation}
\centering
\resizebox{0.35\textwidth}{!}{
$
\left\{
\begin{aligned}
& \Theta_c:=\mathcal{G}(x^{(j)}_{H^{(i)}};\Theta_{\mathcal{G}_s})+\Theta_c \\
& \mathop{\rm min}_{\Theta_{\mathcal{G}_s}} \mathcal{L}_{{\rm{\moduleBbrief{}}}}=\sum_{i=1}^{\mathcal{N}_d} \sum_{j=1}^{ \mathcal{N}_{R^{(i)}}} l_{ce} (y^{(j)}_{H^{(i)}}, \Omega(x^{(j)}_{H^{(i)}};\Theta_b); \Theta_c).
\end{aligned}
\right.
$}
\label{eq:pam_loss}
\end{equation}

Here, we make a basic assumption in our model.
\begin{assumption}
\label{asm:lipschitz}
    \begin{sloppypar}
    (Continuity) Assume that the model satisfies the Lipschitz continuity condition (i.e., $\Vert f(x;{\rm{Clip}}(\Theta^{n}_\triangle)+ \Theta^{n})-f(x;\Theta^{n})\Vert\leq p\Vert({\rm{Clip}}(\Theta^{n}_\triangle)+ \Theta^{n})-\Theta^{n} \Vert=p\Vert \rm{Clip}(\Theta^{n}_\triangle)\Vert$, where $f$ denote the forward propagation of \modelDevice{}, $p$ stands for Lipschitz constant), we achieve this goal by applying the clip operation to $\Theta^{n}_\triangle$. 
    \end{sloppypar}
\end{assumption}

Then, we can have the following lemma:
\begin{lemma}
\label{lemma:bound}
Based on the Assumption~\ref{asm:lipschitz}, by constraining the parameter shift range to \([-\mathcal{T}, \mathcal{T}]\), the generalization error bound is reduced by:
\begin{equation}
\Delta R = 2 \cdot \left( \frac{K - K_{\Delta W}}{\sqrt{n}} \left( \sqrt{m_\mathcal{G}} \right) \right).
\end{equation}
    In the above equation, \(K\) and \(K_{\Delta W}\) represent the constant terms of model complexity under different conditions. Adding parameter constraints reduces model complexity, therefore \(K > K_{\Delta W}\). Since \(K > K_{\Delta W} > 0\), \(n > 0\), and \(m_\mathcal{G} > 0\), it follows that \(\Delta R > 0\). Therefore, our \method{} has a tighter generalization error bound.
\end{lemma}
The proof of Lemma~\ref{lemma:bound} can be found in Appendix.

\subsection{Multi-Prototype \method{}}
\begin{figure}[!h]
    \begin{center}
        \vspace{-0.18cm}
        \includegraphics[width=0.38\textwidth]{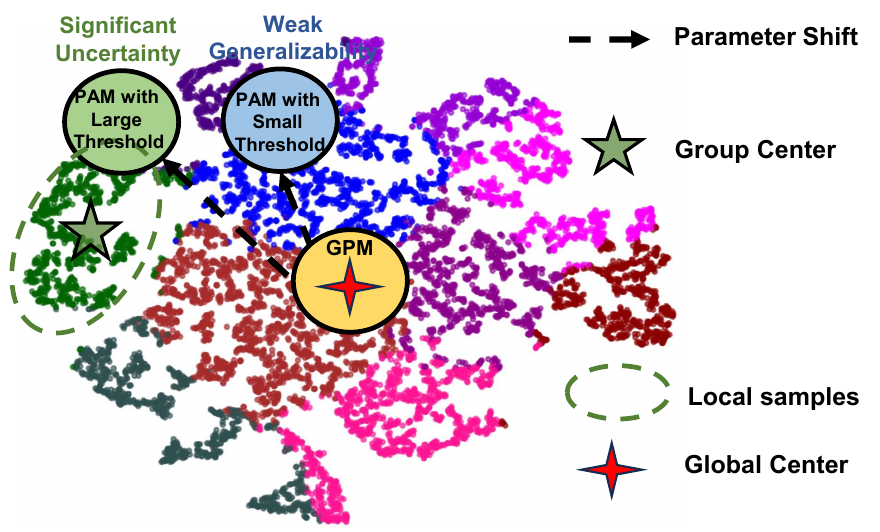}
        \vspace{-0.35cm}        
    \end{center}
    \caption{t-SNE visualization of the data distribution and \modelGradient{} threshold.}
    \label{fig:vis_method}
\vspace{-0.18cm}
\end{figure}

The \method{} framework primarily addresses the discrepancy between the global data distribution and various data distributions. It adapts the Global \moduleA{} (\moduleAbrief{}) with an adaptive \modelGradient{} to diverse devices, thereby delivering the personalization of the local device model.
Figure~\ref{fig:vis_method} illustrates the main intuition behind our \method{} framework. It generalizes the effectiveness of the \moduleAbrief{} on central features of global data to features that are farther away, based on the generated \modelGradient{} for the Primary Adaptive Model (\modelDevice{}).
However, we empirically found that the unconstrained \moduleB{} falls in struggles to generate the appropriate \modelGradient{} for device model personalization. In detail, there is a trade-off of the threshold of generated \modelGradient{}, larger and smaller thresholds respectively introduce significant uncertainty and weak generalizability for effective device model personalization.

\begin{itemize}[itemsep=0pt, topsep=-2pt, leftmargin=*]
\item \textbf{Small Threshold.} Setting a small threshold for \modelGradient{}, for instance, [-0.1, 0.1], can excessively constrain the generated parameter editing matrix. This constraint makes it difficult for the \modelDevice{} to effectively generalize to a large number of samples that are significantly distant from the global data center. As a result, the adaptive model exhibits relatively weak generalizability. 
\item \textbf{Large Threshold.} When a large threshold is set for \modelGradient{}, for instance, [-5.0, 5.0], the adaptive model can theoretically adapt to any data distribution. However, if the local samples are sparse in the data semantic space, a large threshold may introduce significant uncertainty and instability to the model, consequently reducing the device model's performance. 
\end{itemize}

Clearly, the aforementioned stability and generalization ability are contradictory, and it is difficult to achieve when there is only one global \moduleAbrief{} and one parameter change generator. To address these challenges, we propose Multi-Prototype \method{}, which controls the generated parameter changes within a reasonable range by making the \moduleAbrief{}s closer to the samples, to better balance generalization ability and stability. To implement Multi-Prototype \method{}, we propose data partitioning to train multiple \moduleAbrief{}s and \moduleBbrief{}s, and model allocation to match the most suitable \moduleAbrief{} and \moduleBbrief{} to the data during the inference stage.

\noindent\textbf{Initial Partition.}  Assuming the single-prototype \method{} framework is well-trained, we use $\mathcal{M}^{(0)}_\mathcal{G}$ to generate \modelGradient{} for each historical data based on the global \moduleAbrief{} $\mathcal{M}^{(0)}_\mathcal{F}$, so that we can get the \modelGradient{} set $\{G^{(i)}\}_{i=1}^{\mathcal{N}_H}$ corresponding to the historical dataset $\mathcal{D}_H$. 
\begin{equation}
\begin{split}
(\{\mathcal{X}_{H^{(i)}}\}_{i=1}^{\mathcal{N}_H};\{\Theta_{\mathcal{G}_s}, \mathcal{M}^{0}_{\mathcal{F}}), \quad
\{C^{(i)}\}_{i=1}^{\mathcal{N}_H}={\rm{Cluster}}\{G^{(i)}\}_{i=1}^{\mathcal{N}_H}.
\end{split}
\end{equation}
Then, we cluster $\{G^{(i)}\}_{i=1}^{\mathcal{N}_H}$ and obtain the group $C^{(i)}$ a sample $\mathcal{D}_H^{(i)}$ belongs to.

\noindent\textbf{Obtain Group Models.}
After the Initial Partition, devices have been grouped. Then we fine-tune Global \moduleAbrief{} $M^{(0)}_\mathcal{F}$ to $\{M^{(j)}_\mathcal{F}\}_{j=1}^{\mathcal{N}_\mathcal{M}}$. Then, we fine-tune Global \moduleBbrief{} $\mathcal{M}^{(0)}_{\mathcal{G}}$ from $\{\mathcal{M}^{(j)}_{\mathcal{G}}\}_{j=1}^{\mathcal{N}_\mathcal{M}}$-based to $\{\mathcal{M}_\mathcal{G}^{(i)}\}_{i=1}^{\mathcal{N}_\mathcal{M}}$-based.
\begin{equation}
\begin{split}
\resizebox{0.28\textwidth}{!}{
$
\mathcal{M}^{(0)}_{\mathcal{G}} 
\stackrel[\text{Partition }\{C^{(j)}\}_{j=1}^{\mathcal{N}_{\mathcal{M}}}]{\text{Sample }\{\mathcal{X}_{H^{(i)}}\}_{i=1}^{\mathcal{N}_H}}
{\longlonglongrightarrows} \{\mathcal{M}^{(j)}_{\mathcal{G}}\}_{j=1}^{\mathcal{N}_\mathcal{M}},  
\quad
$}
\\
\resizebox{0.28\textwidth}{!}{
$
\mathcal{M}^{(0)}_{\mathcal{F}} 
\stackrel[\text{Partition }\{C^{(j)}\}_{j=1}^{\mathcal{N}_{\mathcal{M}}}]{\text{Sample }\{\mathcal{X}_{H^{(i)}}\}_{i=1}^{\mathcal{N}_H}}
{\longlonglongrightarrows} \{\mathcal{M}^{(j)}_{\mathcal{F}}\}_{j=1}^{\mathcal{N}_\mathcal{M}}.
$}
\end{split}
\end{equation}

\noindent\textbf{Dynamic Assignment.}
\label{subsec:dynamic_partition}
During inference, we use the features of real-time data to obtain a \modelGradient{} on each Group \moduleAbrief{} $\mathcal{M}^{(j)}_{\mathcal{G}}$, and finally obtain a \modelGradient{} set $\{G\}_{j=1}^{\mathcal{N}_\mathcal{M}}$. Then we find the group index $j$ that minimizes the \modelGradient{}, that is, find which Group \moduleAbrief{} $\mathcal{M}^{(j)}_{F}$, the \modelGradient{} is closest to. 
\begin{equation}
\{G^{(j)}\}_{j=1}^{\mathcal{N}_\mathcal{M}}=\{\mathcal{M}^{(j)}_{\mathcal{G}}
(\mathcal{X}_{R^{(i)}};\Theta^{(j)}_{\mathcal{G}}, \mathcal{M}^{(j)}_\mathcal{F})\}^{\mathcal{N}_\mathcal{M}}_{j=1}, 
\end{equation}

Then we choose to use the \modelGradient{} generated by the \moduleBbrief{} of the $i^{th}$ group based on the Group \moduleAbrief{} $\mathcal{M}_\mathcal{F}^{(i)}$ of the $i^{th}$ group, and aggregate the \moduleAbrief{} $\mathcal{M}^{(i)}$ of the $i^{th}$ group into the classifier parameter $\Theta_p$ of the $\mathcal{M}_\mathcal{F}$ and send it to the device.
\begin{equation}
 j^{'}=\arg\min_{j}\{G^{(j)}\}_{j=1}^{\mathcal{N}_\mathcal{M}}, ~~{\rm{Assign}}~\{\mathcal{M}_\mathcal{G}^{(j^{'})}, \mathcal{M}_\mathcal{F}^{(j^{'})}\}
\end{equation}

\begin{table*}[!h]
\setlength{\tabcolsep}{1.5pt}
  \caption{
  Performance comparison of \method{} and baselines on recommendation task 
  (\texttt{Beauty}, \texttt{Electronic}, \texttt{Music}, \texttt{MovieLens}. 
  We mark the results as the \colorbox{best}{best}, \colorbox{second}{second best}.}
\label{tab:main_table}  
\vspace{-0.3cm}
\centering
\renewcommand{\arraystretch}{0.92}
 \resizebox{0.98\textwidth}{!}{
{
    \begin{tabular}{c|c|c|c|c|c|c|c|c|c|c|c|c}
    \toprule[2pt]

    \multirow{2}{*}{\textbf{Baselines}}&\multirow{2}{*}{\textbf{\makecell{\methodsName{}\\Methods}}} & \multicolumn{5}{c|}{\texttt{Electronic Dataset}} & \multicolumn{5}{c|}{\texttt{Beauty Dataset}} & \multirow{2}{*}{\textbf{\makecell{Time\\Consumption} $\downarrow$ }} \\
     \cline{3-7}\cline{8-12}

    & &\textbf{AUC} $\uparrow$ & \textbf{NDCG@5} $\uparrow$ & \textbf{HR@5} $\uparrow$ & \textbf{NDCG@10}  $\uparrow$ & \textbf{HR@10} $\uparrow$ & \textbf{AUC} $\uparrow$ & \textbf{NDCG@5} $\uparrow$ & \textbf{HR@5} $\uparrow$ & \textbf{NDCG@10}  $\uparrow$ & \textbf{HR@10} $\uparrow$ & \\
        \midrule
    \midrule

\multirow{7}{*}{GRU4Rec}& - & 0.7527 & 	0.2468 & 	0.3566 & 	0.2881 & 	0.4863 	 & {0.6761} & \colorbox{second}{0.2103} & \colorbox{second}{0.3156} & \colorbox{second}{0.2483} & \colorbox{second}{0.4351}  & 0\\ \cline{2-7} \cline{9-12} \cline{8-8} \cline{13-13}
    \multirow{7}{*} & TTA & 0.7542 &	0.2449	& 0.3553	& 0.2858	& 0.4815	& 		\colorbox{second}{0.6764} &	0.1985 &	0.3056	& 0.2393 &	0.4311 & 2.07s \\ \cline{2-7} \cline{9-12} \cline{8-8} \cline{12-13}
    \multirow{7}{*} & Fine-tune & 0.7554 & 	0.2482 & 	0.3582 & 	0.2889 & 	0.4842 	 & 0.6712 & 0.1826 & 0.2888 & 0.2272 & 0.4264  & 62.63s \\ \cline{2-7} \cline{9-12} \cline{8-8} \cline{13-13}
     \multirow{7}{*} & DCCL & 0.7598 & 0.2501 & 0.3601 & 0.2898 & 0.4854 & 0.6676 & 0.1842 & 0.2901 & 0.2289 & 0.3874 & 882.4ms \\ \cline{2-7} \cline{9-12} \cline{8-8} \cline{12-13}
    \multirow{7}{*} & DUET & 0.7886 & 0.2976 & 0.4190 & 0.3358 & 0.5371 & 0.6652 & 0.1968 & 0.2892 & 0.2262 & 0.3936 & 10.78ms \\ \cline{2-7} \cline{9-12} \cline{8-8} \cline{12-13}
    \multirow{7}{*} & \cellcolor{blue!5}\method{} (s) & \cellcolor{blue!5}\colorbox{second}{0.7895} & 	\cellcolor{blue!5}\colorbox{second}{0.2976} & 	\cellcolor{blue!5}\colorbox{second}{0.4204} & 	\cellcolor{blue!5}\colorbox{second}{0.3363} & 	\cellcolor{blue!5}\colorbox{second}{0.5398} 	 & \cellcolor{blue!5}0.6668 & \cellcolor{blue!5}0.2013 & \cellcolor{blue!5}0.2895 & \cellcolor{blue!5}0.2329 & \cellcolor{blue!5}0.3942  & \multirow{2}{*}{10.78ms}\\ \cline{2-7} \cline{8-12}
    
    \multirow{7}{*}& \cellcolor{blue!5}\method{} (m) & \cellcolor{blue!5}\colorbox{best}{0.7945} & 	\cellcolor{blue!5}\colorbox{best}{0.3022} & 	\cellcolor{blue!5}\colorbox{best}{0.4233} & 	\cellcolor{blue!5}\colorbox{best}{0.3407} & 	\cellcolor{blue!5}\colorbox{best}{0.5452} & 		\cellcolor{blue!5}\colorbox{best}{0.6785} & \cellcolor{blue!5}\colorbox{best}{0.2255} & \cellcolor{blue!5}\colorbox{best}{0.3256} & \cellcolor{blue!5}\colorbox{best}{0.2541} & \cellcolor{blue!5}\colorbox{best}{0.4400} & \\ \cline{2-7} \cline{8-13}
    
    \midrule[1pt]

\multirow{7}{*}{SASRec}& - & 0.7750 & 	0.2743 & 0.3931 	& 0.3142 & 	0.5162 	& 0.6588 & 0.1672 & 0.2681 & 0.2060 & 0.3892 & 0\\ \cline{2-7} \cline{9-12} \cline{8-8} \cline{13-13}

    \multirow{7}{*} & TTA & 0.7762 &	0.2726 &	0.3903 &	0.3132 &	0.5157	& 		0.6572 &	0.1616 &	0.2587 &	0.2020 &	0.3839 & 2.96s \\ \cline{2-7} \cline{9-12} \cline{8-8} \cline{12-13}
    \multirow{7}{*} & Fine-tune & 0.7554 & 	0.2482 & 	0.3582 & 	0.2889 & 	0.4842 	 & 0.6712 & 0.1826 & 0.2888 & 0.2272 & \colorbox{second}{0.4264}  & 
 89.21s \\ \cline{2-7} \cline{9-12} \cline{8-8} \cline{13-13}

   \multirow{7}{*} & DCCL & 0.7754 & 0.2738 & 0.3919 & 0.3143 & 0.5167 & 0.6598 & 0.1587 & 0.2562 & 0.1998 & 0.3867 & 882.4ms \\ \cline{2-7} \cline{9-12} \cline{8-8} \cline{12-13}
   \multirow{7}{*} & DUET & 0.7841 & 0.2934 & 0.4151 & 0.3310 & 0.5311 & 0.6695 & 0.2064 & 0.2982 & \colorbox{second}{0.2354} & 0.4040 & 10.78ms \\ \cline{2-7} \cline{9-12} \cline{8-8} \cline{12-13}
    
    \multirow{7}{*} & \cellcolor{blue!5}\method{} (s) & \cellcolor{blue!5}\colorbox{second}{0.7805} & 	\cellcolor{blue!5}\colorbox{second}{0.2898} & \cellcolor{blue!5}\colorbox{second}{0.4102} 	& \cellcolor{blue!5}\colorbox{second}{0.3278} & 	\cellcolor{blue!5}\colorbox{second}{0.5296} 		& \cellcolor{blue!5}\colorbox{second}{0.6729} & \cellcolor{blue!5}\colorbox{second}{0.2029} & \cellcolor{blue!5}\colorbox{second}{0.2999} & \cellcolor{blue!5}\colorbox{second}{0.2331} & \cellcolor{blue!5}0.3976  & \multirow{2}{*}{10.78ms}\\ \cline{2-7} \cline{8-12}
    
    \multirow{7}{*}& \cellcolor{blue!5}\method{} (m) & \cellcolor{blue!5}\colorbox{best}{0.7934} & 	\cellcolor{blue!5}\colorbox{best}{0.2991} & \cellcolor{blue!5}\colorbox{best}{0.4242} 	& \cellcolor{blue!5}\colorbox{best}{0.3367} & 	\cellcolor{blue!5}\colorbox{best}{0.5430} 	& \cellcolor{blue!5}\colorbox{best}{0.6820} & \cellcolor{blue!5}\colorbox{best}{0.2241} & \cellcolor{blue!5}\colorbox{best}{0.3304} & \cellcolor{blue!5}\colorbox{best}{0.2592} & \cellcolor{blue!5}\colorbox{best}{0.4407}   & \\
    \bottomrule
    \bottomrule

    \multirow{2}{*}{\textbf{Baselines}}&\multirow{2}{*}{\textbf{\makecell{\methodsName{}\\Methods}}} & \multicolumn{5}{c|}{\texttt{Music Dataset}} & \multicolumn{5}{c|}{\texttt{MovieLens Dataset}} & \multirow{2}{*}{\textbf{\makecell{Time\\Consumption} $\downarrow$ }} \\
     \cline{3-7}\cline{8-12}

    & &\textbf{AUC} $\uparrow$ & \textbf{NDCG@5} $\uparrow$ & \textbf{HR@5} $\uparrow$ & \textbf{NDCG@10}  $\uparrow$ & \textbf{HR@10} $\uparrow$ & \textbf{AUC} $\uparrow$ & \textbf{NDCG@5} $\uparrow$ & \textbf{HR@5} $\uparrow$ & \textbf{NDCG@10}  $\uparrow$ & \textbf{HR@10} $\uparrow$ & \\
    \midrule
    \midrule

\multirow{7}{*}{GRU4Rec}& - & \colorbox{second}{0.8653} 	& 0.3802 	& \colorbox{second}{0.5105} 	& 0.4219 	& \colorbox{second}{0.6395} 		& 		0.9093 & 	0.4903 & 	0.6491 & 	0.5274 & 	0.7619  & 0\\ \cline{2-7} \cline{9-12} \cline{8-8} \cline{13-13}
    \multirow{7}{*} & TTA & 0.8652 &	0.3777 &	0.5100 &	0.4196 &	0.6400 &	0.9100 &	0.4953 &	0.6508 &	0.5323 &	0.7643 & 2.07s \\ \cline{2-7} \cline{9-12} \cline{8-8} \cline{12-13}
    \multirow{7}{*} & Fine-tune & 0.8660 &	0.3752 &	0.5067 &	0.4180 &	0.6390 &	0.9114 &	0.4942 &	0.6515 &	0.5314 & 0.7654 & 62.63s \\ \cline{2-7} \cline{9-12} \cline{8-8} \cline{13-13}
    \multirow{7}{*} &  DCCL & 0.8632 & 0.3748 & 0.5072 & 0.4171 & 0.6374 & 	0.9089 & 0.4938 & 0.6521 & 0.5302 & 0.7657 & 882.4ms\\ \cline{2-7} \cline{9-12} \cline{8-8} \cline{12-13}
    \multirow{7}{*} & DUET & 0.8645 & 0.3795 & 0.5068 & 0.4217 & 0.6385	& \colorbox{second}{0.9103} & \colorbox{second}{0.5009} & \colorbox{second}{0.6562} & \colorbox{second}{0.5365} & \colorbox{second}{0.7668} & 10.78ms \\ \cline{2-7} \cline{9-12} \cline{8-8} \cline{12-13}
    
    \multirow{7}{*} & \cellcolor{blue!5}\method{} (s) & \cellcolor{blue!5}0.8644 	& \cellcolor{blue!5}\colorbox{second}{0.3812} 	& \cellcolor{blue!5}0.5089 	& \cellcolor{blue!5}\colorbox{second}{0.4220} 	& \cellcolor{blue!5}0.6384 		& 		\cellcolor{blue!5}0.9102 &  \cellcolor{blue!5}0.4948 & 	\cellcolor{blue!5}0.6522 &  \cellcolor{blue!5}0.5311 & 	\cellcolor{blue!5}0.7640  & \multirow{2}{*}{10.78ms}\\ \cline{2-7} \cline{8-12}
    
    \multirow{7}{*}& \cellcolor{blue!5}\method{} (m) & \cellcolor{blue!5}\colorbox{best}{0.8685} 	& \cellcolor{blue!5}\colorbox{best}{0.3899} 	& \cellcolor{blue!5}\colorbox{best}{0.5182} 	& \cellcolor{blue!5}\colorbox{best}{0.4320} 	& \cellcolor{blue!5}\colorbox{best}{0.6491} 		& 		\cellcolor{blue!5}\colorbox{best}{0.9111} & 	\cellcolor{blue!5}\colorbox{best}{0.5036} & 	\cellcolor{blue!5}\colorbox{best}{0.6601} & 	\cellcolor{blue!5}\colorbox{best}{0.5389} & 	\cellcolor{blue!5}\colorbox{best}{0.7684}  & \\ \cline{2-7} \cline{8-13}

    \midrule[1pt]

\multirow{7}{*}{SASRec}& - & 0.8633 & 	0.3808 & 	0.5115 & 0.4218 & 	0.6395 & 	0.9080 & 	0.4938 & 	0.6505 & 	0.5304 & 	0.7636   & 0\\ \cline{2-7} \cline{9-12} \cline{8-8} \cline{13-13}
    \multirow{7}{*}  & TTA & 0.8623 &	0.3773 &	0.5102 &	0.4191 &	0.6397 &	0.9083 &	0.4951 &	0.6508 &	0.5323 &	0.7652 & 2.96s \\ \cline{2-7} \cline{9-12} \cline{8-8} \cline{12-13}
    
    \multirow{7}{*} & Fine-tune & 0.8644 & 	0.3755 & 	0.5086 & 	0.4180 &  0.6403 & 	\colorbox{best}{0.9104} & 	\colorbox{second}{0.4993} & 	\colorbox{second}{0.6563} & 	\colorbox{second}{0.5352} & 	\colorbox{second}{0.7668} & 89.21s \\ \cline{2-7} \cline{9-12} \cline{8-8} \cline{13-13}

    \multirow{7}{*} & DCCL & 0.8642 & 0.3748 & 0.5092 & 0.4168 & 0.6411 &  0.9097 & 0.4957 & 0.6486 & 0.5207 & 0.7631 & 882.4ms\\ \cline{2-7} \cline{9-12} \cline{8-8} \cline{12-13}
    \multirow{7}{*} & DUET & 0.8654 & 0.3857 & 0.5120 & 0.4268 & 0.6408 & 0.9102 & 0.4968 & 0.6548 & 0.5336 & 0.7667 & 10.78ms \\ \cline{2-7} \cline{9-12} \cline{8-8} \cline{12-13}
    \multirow{7}{*} & \cellcolor{blue!5}\method{} (s) & \cellcolor{blue!5}\colorbox{best}{0.8682} & 	\cellcolor{blue!5}\colorbox{second}{0.3852} & 	\cellcolor{blue!5}\colorbox{second}{0.5144} & \cellcolor{blue!5}\colorbox{second}{0.4274} & 	\cellcolor{blue!5}\colorbox{best}{0.6443} & 	\cellcolor{blue!5}0.9100 & 	\cellcolor{blue!5}0.4956 & 	\cellcolor{blue!5}0.6527 & 	\cellcolor{blue!5}0.5325 & 	\cellcolor{blue!5}0.7652  & \multirow{2}{*}{10.78ms}\\ \cline{2-7} \cline{8-12}
 
    \multirow{7}{*}& \cellcolor{blue!5}\method{} (m) & \cellcolor{blue!5}\colorbox{second}{0.8668} & 	\cellcolor{blue!5}\colorbox{best}{0.3905} & 	\cellcolor{blue!5}\colorbox{best}{0.5192} & \cellcolor{blue!5}\colorbox{best}{0.4308} & 	\cellcolor{blue!5}\colorbox{best}{0.6443} & 	\cellcolor{blue!5}\colorbox{best}{0.9104} & 	\cellcolor{blue!5}\colorbox{best}{0.5031} & 	\cellcolor{blue!5}\colorbox{best}{0.6597} & 	\cellcolor{blue!5}\colorbox{best}{0.5381} & 	\cellcolor{blue!5}\colorbox{best}{0.7669} & \\ 
    
    \bottomrule[2pt]

    \end{tabular}
    }
}
\vspace{-0.2cm}
\end{table*}

\noindent\textbf{Cross-Layer Knowledge Transfer.} 
\label{subsec:knowledge_transfer}
Through the above methods, we can get Multi-Prototype \method{}. However, Multi-Prototype \method{} has the weakness of partition inconsistency.
The partition inconsistency can be mainly described as: assuming that the Adaptive layer of \modelDevice{} has a total of $N$ layers, \moduleBbrief{} needs to generate the \modelGradient{}s of the $N$ layer neural network based on the data. However, there is no information transfer between the $N$ \moduleBbrief{} branches used to generate the $N$ layer \modelGradient{}s. The lack of information transfer between branches causes \moduleBbrief{} to generate \modelGradient{}s for $N$ linear layers that are independent of each other instead of $N$ layers of interconnected neural networks.
This results in inconsistent partition results when the same sample is grouped using \modelGradient{} generated by different \moduleBbrief{} branches, that is, the same sample is divided into different groups. 

In order to improve the partition consistency, we constrain the feature $\boldsymbol{e}^{(j)}_{H}$ used to generate the \modelGradient{} of the $j^{th}$ layer, so that the feature $\boldsymbol{e}^{(j+1)}_{H}$ used for the $(j+1)^{th}$ layer \modelGradient{} depends on $\boldsymbol{e}^{(j)}_{H}$. In this way, we constrain the \modelGradient{} produced by the same sample to be in the same space.
\begin{equation}
\Omega_{seq}(\{\boldsymbol{e}^{n}_{R^{(i)}}\}_{n=1}^{\mathcal{N}_l})\longrightarrow \{\boldsymbol{e}^{n}_{R^{(i)}}\}_{n=1}^{\mathcal{N}_l}
\end{equation}
where $\Omega_{seq}$ represents a sequential modeling module.

\section{Experiments}
\label{sec:experiments}
We conduct a range of experiments on tasks of two modality (vision and recommendation). Part of the experimental setup, results and analysis in the Appendix.

\begin{table*}[!h]
\setlength{\tabcolsep}{2.5pt}
  \caption{
  Performance comparison of \method{} and baselines on vision task (\texttt{CK+}). 
  We mark the results as the \colorbox{best}{best}, \colorbox{second}{second best}.
  }
\vspace{-0.3cm}
\label{tab:main_table_cv}
\centering
\renewcommand{\arraystretch}{0.92}
\resizebox{\textwidth}{!}{
\begin{tabular}{c|c|c|c|c|c|c|c|c|c|c|c}
\toprule[2pt]
 &  & \multicolumn{3}{c|}{\texttt{CK+ Dataset}} & \multirow{2}{*}{\textbf{\makecell{Time\\Consumption} $\downarrow$}}  &  &  & \multicolumn{3}{c|}{\texttt{CK+ Dataset}} & \multirow{2}{*}{\textbf{\makecell{Time\\Consumption} $\downarrow$}}\\ 
 \cline{3-5} \cline{9-11}
\multirow{-2}{*}{\textbf{\textbf{Baselines}}} & \multirow{-2}{*}{\textbf{\makecell{\methodsName{}\\Methods}}} & \multicolumn{1}{c|}{\textbf{ACC}$\uparrow$} & \multicolumn{1}{c|}{\textbf{AUC}$\uparrow$} & \multicolumn{1}{c|}{\textbf{{ACC@3}$\uparrow$}} & & \multirow{-2}{*}{\textbf{Baselines}} & \multirow{-2}{*}{\textbf{\makecell{\methodsName{}\\Methods}}} & \multicolumn{1}{c|}{\textbf{ACC}$\uparrow$} & \multicolumn{1}{c|}{\textbf{AUC}$\uparrow$} & \multicolumn{1}{c|}{\textbf{{ACC@3}$\uparrow$}} &  \\ 
\midrule \midrule
 & - & 57.58 & 61.04 & 81.44 & 0 & & - & 68.69 & 76.16 & 86.62 & 0  \\ \cline{2-6} \cline{8-12} 
  & TTA & 70.69 & 75.73 & 83.15 & 3.29s & & TTA & 70.56 & 78.79 & 87.61 & 3.91s \\ \cline{2-6} \cline{8-12} 
 & Fine-tune & \colorbox{second}{71.72} & 78.26 & 85.35 & 16.49s & & Fine-tune & \colorbox{second}{71.92} & \colorbox{second}{79.99} & \colorbox{second}{88.13} & 19.50s \\ \cline{2-6} \cline{8-12} 
 & DUET & 67.68 & \colorbox{second}{78.64} & \colorbox{second}{86.87} & 15.18ms & & DUET & 71.47 & 79.03 & 88.12 & 15.18ms \\ \cline{2-6} \cline{8-12} 
 \multirow{-5}{*}{MobileNetV3-Small} & \cellcolor{blue!10}\method{} & \cellcolor{blue!10}\colorbox{best}{73.74}& \cellcolor{blue!10}\colorbox{best}{80.06} & \cellcolor{blue!10}\colorbox{best}{87.88} & 15.18ms & \multirow{-5}{*}{MobileNetV3-Large} & \cellcolor{blue!10}\method{} & \cellcolor{blue!10}\colorbox{best}{74.76} & \cellcolor{blue!10}\colorbox{best}{81.01} & \cellcolor{blue!10}\colorbox{best}{89.92} & 18.83ms \\ 
\bottomrule[2pt]
\end{tabular}
}
\vspace{-0.15cm}
\end{table*}

\begin{table*}[!h]
\setlength{\tabcolsep}{2.5pt}
  \caption{
  Effectiveness of Data Partition.
  } 
\vspace{-0.3cm}
  \label{tab:group_finetuning}
  \centering

   \renewcommand{\arraystretch}{0.92}
 \resizebox{\textwidth}{!}{
    \begin{tabular}{c|c|c|c|c|c|c|c|c|c|c|c|c}
    \toprule[2pt]
     \multirow{2}{*}{\textbf{Baselines}}&\multirow{2}{*}{\textbf{\makecell{\methodsName{}\\Methods}}} & \multirow{2}{*}{\textbf{\makecell{Group\\Numbers}}} & \multicolumn{5}{c|}{\texttt{Beauty Dataset}} & \multicolumn{5}{c}{\texttt{Electronic Dataset}}  \\
     \cline{4-7}\cline{8-13}
    & & & \textbf{AUC} $\uparrow$ & \textbf{NDCG@5} $\uparrow$ & \textbf{HR@5} $\uparrow$ & \textbf{NDCG@10}  $\uparrow$ & \textbf{HR@10} $\uparrow$ & \textbf{AUC} $\uparrow$ & \textbf{NDCG@5} $\uparrow$  & \textbf{HR@5} $\uparrow$ & \textbf{NDCG@10} $\uparrow$ & \textbf{HR@10} $\uparrow$ \\ 
    \midrule \midrule
\multirow{10}{*}{SASRec}& \cellcolor{gray!0}-  & \cellcolor{gray!0}- & \cellcolor{gray!0}0.6588 & \cellcolor{gray!0}0.1672 & \cellcolor{gray!0}0.2681 & \cellcolor{gray!0}0.2060 & \cellcolor{gray!0}0.3892 & \cellcolor{gray!0}0.7750 & \cellcolor{gray!0}0.2743 & \cellcolor{gray!0}0.3931 & \cellcolor{gray!0}0.3142 & \cellcolor{gray!0}0.5162 \\ \cline{2-7} \cline{9-12} \cline{8-8} \cline{13-13}
\multirow{10}{*} & Fine-tuning  & \cellcolor{gray!0}- & \cellcolor{gray!0}0.6565 & \cellcolor{gray!0}0.1558 & \cellcolor{gray!0}0.2520 & \cellcolor{gray!0}0.1983 & \cellcolor{gray!0}0.3839 & \cellcolor{gray!0}0.7754 & \cellcolor{gray!0}0.2738 & \cellcolor{gray!0}0.3919 & \cellcolor{gray!0}0.3143 & \cellcolor{gray!0}0.5167 \\ \cline{2-7} \cline{9-12} \cline{8-8} \cline{13-13}
\multirow{10}{*} & \multirow{4}{*}{\makecell{Group\\Fine-tuning}}  & \cellcolor{blue!2}2 & \cellcolor{blue!2}0.6588 & \cellcolor{blue!2}0.2066 & \cellcolor{blue!2}0.3082 & \cellcolor{blue!2}0.2353 & \cellcolor{blue!2}0.4003 & \cellcolor{blue!2}0.7641 & \cellcolor{blue!2}0.2541 & \cellcolor{blue!2}0.3653 & \cellcolor{blue!2}0.2936 & \cellcolor{blue!2}0.4877 \\ \cline{3-7} \cline{9-12} \cline{8-8} \cline{13-13}
\multirow{10}{*} &  & \cellcolor{blue!2}3 & \cellcolor{blue!2}0.6645 & \cellcolor{blue!2}0.2139 & \cellcolor{blue!2}0.3146 & \cellcolor{blue!2}0.2441 & \cellcolor{blue!2}0.4113 & \cellcolor{blue!2}0.7702 & \cellcolor{blue!2}0.2541 & \cellcolor{blue!2}0.3653 & \cellcolor{blue!2}0.2933 & \cellcolor{blue!2}0.4858 \\ \cline{3-7} \cline{9-12} \cline{8-8} \cline{13-13}
\multirow{10}{*} &  & \cellcolor{blue!2}5 & \cellcolor{blue!2}0.6657 & \cellcolor{blue!2}0.2077 & \cellcolor{blue!2}0.3096 & \cellcolor{blue!2}0.2365 & \cellcolor{blue!2}0.4076 & \cellcolor{blue!2}0.7806 & \cellcolor{blue!2}0.2731 & \cellcolor{blue!2}0.3896 & \cellcolor{blue!2}0.3129 & \cellcolor{blue!2}0.5128 \\ \cline{3-7} \cline{9-12} \cline{8-8} \cline{13-13}
\multirow{10}{*} &  & \cellcolor{blue!2}10 & \cellcolor{blue!2}0.6647 & \cellcolor{blue!2}0.2059 & \cellcolor{blue!2}0.3076 & \cellcolor{blue!2}0.2338 & \cellcolor{blue!2}0.4066 & \cellcolor{blue!2}0.7824 & \cellcolor{blue!2}0.2893 & \cellcolor{blue!2}0.4072 & \cellcolor{blue!2}0.3267 & \cellcolor{blue!2}0.5229 \\ \cline{2-7} \cline{9-12} \cline{8-8} \cline{13-13}
\multirow{10}{*} & \multirow{4}{*}{\method{}}  & \cellcolor{blue!5}2 & \cellcolor{blue!5}0.6788 & \cellcolor{blue!5}0.2119 & \cellcolor{blue!5}0.3196 & \cellcolor{blue!5}0.2461 & \cellcolor{blue!5}0.4291 & \cellcolor{blue!5}0.7915 & \cellcolor{blue!5}0.2911 & \cellcolor{blue!5}0.4135 & \cellcolor{blue!5}0.3312 & \cellcolor{blue!5}0.5376 \\ \cline{3-7} \cline{9-12} \cline{8-8} \cline{13-13}
\multirow{10}{*} &  & \cellcolor{blue!5}3 & \cellcolor{blue!5}0.6784 & \cellcolor{blue!5}0.2163 & \cellcolor{blue!5}0.3253 & \cellcolor{blue!5}0.2506 & \cellcolor{blue!5}0.4317 & \cellcolor{blue!5}0.7914 & \cellcolor{blue!5}0.2944 & \cellcolor{blue!5}0.4163 & \cellcolor{blue!5}0.3328 & \cellcolor{blue!5}0.5372 \\ \cline{3-7} \cline{9-12} \cline{8-8} \cline{13-13}
\multirow{10}{*} &  & \cellcolor{blue!5}5 & \cellcolor{blue!5}0.6832 & \cellcolor{blue!5}0.2188 & \cellcolor{blue!5}0.3280 & \cellcolor{blue!5}0.2550 & \cellcolor{blue!5}0.4394 & \cellcolor{blue!5}0.7929 & \cellcolor{blue!5}0.2972 & \cellcolor{blue!5}0.4200 & \cellcolor{blue!5}0.3355 & \cellcolor{blue!5}0.5390  \\ \cline{3-7} \cline{9-12} \cline{8-8} \cline{13-13}
\multirow{10}{*} &  & \cellcolor{blue!5}10 & \cellcolor{blue!5}0.6820 & \cellcolor{blue!5}0.2241 & \cellcolor{blue!5}0.3304 & \cellcolor{blue!5}0.2592 & \cellcolor{blue!5}0.4407 & \cellcolor{blue!5}0.7934 & \cellcolor{blue!5}0.2991 & \cellcolor{blue!5}0.4242 & \cellcolor{blue!5}0.3367 & \cellcolor{blue!5}0.5430 \\  \bottomrule[2pt] 
\end{tabular}
}
\vspace{-0.15cm}
\end{table*}

\subsection{Datasets}
\begin{sloppypar}
\noindent\textbf{Datasets.}  We evaluate \method{} and user modeling baselines on \texttt{Amazon Beauty~(Beauty)}, \texttt{Amazon CDs~(CDs)}, \texttt{Amazon Electronic~(Electronic)}, \texttt{Douban Book~(Book)}, \texttt{Douban Music~(Music)}, \texttt{MovieLens-1M~(MovieLens)}, 6 widely used public real-world sequential user modeling datasets. We evaluate \method{} and vision baselines on \texttt{Extended Cohn-Kanade dataset~(CK+)}, a widely used public image classification dataset. The reason why CIFAR-10, ImageNet and other datasets are not used is because CK+ is a dataset that can reflect changes in user distribution on the device. Due to the more complex data distribution in user modeling task, the distribution shift of on-device data is more pronounced and rapid. Therefore,the experiments are more conducted on user modeling task.
\end{sloppypar}

\begin{sloppypar}
\noindent\textbf{Baselines.}
\label{subsec:experiment_baseline}
\noindent \textit{On-device deep learning models}:
GRU4Rec~\cite{ref:gru4rec} and SASRec~\cite{ref:sasrec} are used in the recommendation, with SASRec as the default baseline. MobileNetV3-large~\cite{ref:mobilenetv3} and MobileNetV3-small~\cite{ref:mobilenetv3} are used in the vision task.  Models like ResNet and BERT4Rec~\cite{ref:bert4rec}, despite superior performance, were excluded due to high resource demands in device deployment. 
\noindent \textit{Retraining-based \methodsName{} methods}:
Fine-tune~\cite{ref:finetune} and Test-time Adaptation(TTA)~\cite{ref:test_time_adaptation} are two widely used methods for improving the model generalizability via retraining.
\noindent \textit{Device-cloud collaboration frameworks}: DCCL~\cite{ref:dccl} and DUET~\cite{ref:duet}. These frameworks were selected due to their similarity in tasks, enabling a meaningful and insightful comparison. DCCL we only compare with it on the recommendation task due to it is designed for recommendation task. 
\end{sloppypar}

\noindent\textbf{Evaluation Metrics.}
In the experiments, we use the widely adopted AUC, HitRate, and NDCG as the metrics to evaluate model performance. 
HitRate and NDCG are both obtained with top-5 and top-10 recommendations.
The details of the datasets (including preprocessing procedure), baselines, and evaluation metrics can be found in the Appendix.

\noindent\textbf{Symbol Description.}
$\uparrow$ means that the larger the value, the better, and $\downarrow$ means that the smaller the value, the better. $\Uparrow$ means upload and $\Downarrow$ means download. \textbf{w.} and \textbf{w/o.} are represented ``with'' and ``without'', respectively.

\subsection{Quantitative Results}
\label{subsubsec:quantitative_results}

\begin{figure*}[!h]
  \centering
\includegraphics[width=0.96\linewidth]{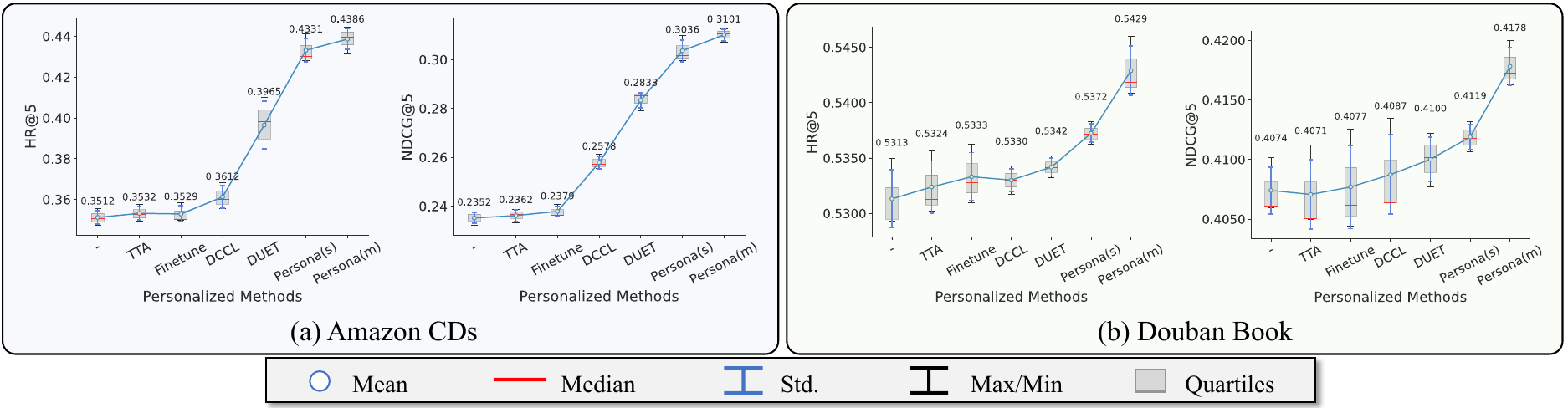}
\vspace{-0.25cm}
  \caption{Performance \emph{w.r.t.} \methodsName{} methods.}
  \label{fig:main_tab}
\vspace{-0.15cm}
\end{figure*}

Table~\ref{tab:main_table} summarizes the quantitative results of our \method{} framework and other \method{} methods on recommendation task. From this table, we have the following findings: 
(1) Almost all \methodsName{} methods can improve the baseline's ("-") performance, which reveals the significance of model generalization on the device.
(2) In all cases, the effect of model fine-tuning~(Fine-tune) and TTA are insignificant. What's more serious is we observe performance degradation in some cases,
(\emph{e.g.}, \{Dataset=\texttt{Beauty}, Model$\in$\{GRU4Rec, SASRec\}, Metric$\in$\{AUC, NDCG, HR\}\}). This phenomenon is reasonable as the fine-tuning model may encounter the over-fitting issue when trained on limited real-time data. Besides, the results also indicate that this \method{} method causes a high delay, which is impractical for applications on the device. 
(3) In all cases, the single-prototype \method{}~(\method{} (s)) and multi-prototype \method{}~(\method{} (m)) both outperform retraining-based \methodsName{} methods and base model by an large margin.
Notably, it enables real-time personalized sequential recommendation with an extremely low time consumption compared with retraining-based \methodsName{} methods and DCCL. Compared with DUET, Persona is much better than DUET in terms of performance, although the time consumption is similar.
(4) In most cases, multi-prototype \method{} performs better than single-prototype \method{}, which shows that model editing based on multi-prototype has better generalization ability than model editing based on global model.
In summary, the aforementioned results strongly demonstrate the effectiveness and generalizability of our proposed \method{}. 
Table~\ref{tab:main_table_cv} summarizes the quantitative results of our \method{} framework and other \methodsName{} methods on vision task. We find the same conclusion as in the recommendation task. Figure~\ref{fig:main_tab} shows the mean error bar of our \method{} framework and other \methodsName{} methods.

\begin{figure*}[!h]
  \centering
\includegraphics[width=0.98\linewidth]{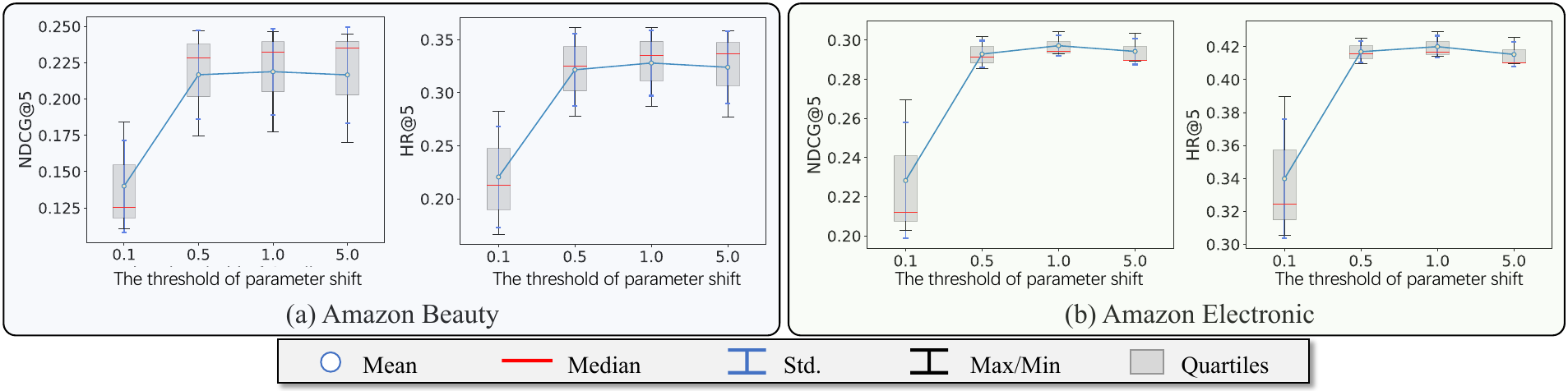}
\vspace{-0.25cm}
  \caption{
  The impact of the threshold of \modelGradient{}.
  }
  \label{fig:grad_thres}
  \vspace{-0.25cm}
\end{figure*}

\subsection{Ablation Study}
\label{subsubsec:effectiveness_evaluation}
This subsection shows ablation study~(\textit{i.e}, the effectiveness of data partition, cross-layer knowledge transfer and dynamic assignment).

\noindent \textbf{Effectiveness of Data Partition.}
We analyzed the effectiveness of Data Partition in Table~\ref{tab:group_finetuning}.
In order to evaluate the effectiveness of data partition, we performed Group Fine-tuning after data partition, and compared it with \method{}, Finetuning, and without \method{} method to evaluate whether it can improve the performance. At the same time, we also set the number of groups $\mathcal{N}_\mathcal{M}\in\{2, 3, 5, 10\}$ to observe the change of performance. As shown in Table~\ref{tab:group_finetuning}, we have the following findings:
(1) If the variable $\mathcal{N}_\mathcal{M}$ is appropriate (such as $\mathcal{N}_\mathcal{M}\in\{3, 5, 10\}$ on the Beauty dataset, $\mathcal{N}_\mathcal{M}\in\{5, 10\}$ on the Electronic dataset), Group Fine-tuning is compared to Finetuning, and without \method{} method, can bring performance improvements, which also shows that the effectiveness our data partition.
(2) On the small dataset Beauty, Group Fine-tuning achieves the optimal performance when $\mathcal{N}_\mathcal{M}=5$. On the larger dataset Electronic, Group Fine-tuning achieves the optimal performance when $\mathcal{N}_\mathcal{M}=10$. This shows that the optimal $\mathcal{N}_\mathcal{M}$ is related to the size of the dataset. 
(3) Under different $\mathcal{N}_\mathcal{M}$ values, \method{} achieves better performance than group fine-tuning, revealing its effectiveness.

\begin{table}[!h]

\setlength{\tabcolsep}{2pt}
\vspace{-0.1cm}
  \caption{
  Effectiveness of Cross-Layer Knowledge Transfer.
  } 
  \vspace{-0.3cm}
  \label{tab:ablation_clkt}
  \centering
 \renewcommand{\arraystretch}{0.96}
 \resizebox{0.475\textwidth}{!}{
    \begin{tabular}{c|c|c|c|c|c|c}
    \toprule[2pt]
 \multirow{2}{*}{\textbf{\makecell{Partition\\Methods}}} & \multirow{2}{*}{\textbf{\makecell{Group\\Numbers}}} & \multicolumn{5}{c}{\texttt{Beauty Dataset}} \\
     
\cline{3-7}
    \multirow{2}{*} & \multirow{2}{*} & \textbf{AUC} $\uparrow$ & \textbf{NDCG@5} $\uparrow$ & \textbf{HR@5} $\uparrow$ & \textbf{NDCG@10}  $\uparrow$ & \textbf{HR@10} $\uparrow$ \\
   
    \midrule
    \midrule

 \multirow{4}{*}{\makecell{\textbf{w/o.}\\CLKT}}  & \cellcolor{blue!2}2 & \cellcolor{blue!2}0.6654 & \cellcolor{blue!2}0.1756 & \cellcolor{blue!2}0.2754 & \cellcolor{blue!2}0.2162 & \cellcolor{blue!2}0.4016 \\  \cline{2-7}
  & \cellcolor{blue!2}3 & \cellcolor{blue!2}0.6647 & \cellcolor{blue!2}0.1735 & \cellcolor{blue!2}0.2734 & \cellcolor{blue!2}0.2149 & \cellcolor{blue!2}0.4013 \\  \cline{2-7}
  & \cellcolor{blue!2}5 & \cellcolor{blue!2}0.6638 & \cellcolor{blue!2}0.1705 & \cellcolor{blue!2}0.2694 & \cellcolor{blue!2}0.2116 & \cellcolor{blue!2}0.3963 \\  \cline{2-7}
  & \cellcolor{blue!2}10 & \cellcolor{blue!2}0.6639 & \cellcolor{blue!2}0.1702 & \cellcolor{blue!2}0.2664 & \cellcolor{blue!2}0.2115 & \cellcolor{blue!2}0.3956 \\  \cline{1-7}

 \multirow{4}{*}{\makecell{\textbf{w.}\\CLKT}}  & \cellcolor{blue!5}2 & \cellcolor{blue!5}0.6685 & \cellcolor{blue!5}0.2184 & \cellcolor{blue!5}0.3193 & \cellcolor{blue!5}0.2499 & \cellcolor{blue!5}0.4153 \\  \cline{2-7}
  & \cellcolor{blue!5}3 & \cellcolor{blue!5}0.6690 & \cellcolor{blue!5}0.2225 & \cellcolor{blue!5}0.3216 & \cellcolor{blue!5}0.2511 & \cellcolor{blue!5}0.4160 \\  \cline{2-7}
  & \cellcolor{blue!5}5 & \cellcolor{blue!5}0.6688 & \cellcolor{blue!5}0.2231 & \cellcolor{blue!5}0.3266 & \cellcolor{blue!5}0.2520 & \cellcolor{blue!5}0.4207 \\  \cline{2-7}
  & \cellcolor{blue!5}10 & \cellcolor{blue!5}0.6677 & \cellcolor{blue!5}0.2228 & \cellcolor{blue!5}0.3243 & \cellcolor{blue!5}0.2527 & \cellcolor{blue!5}0.4173 \\
\bottomrule[2pt]
\end{tabular}
}
\vspace{-0.1cm}
\end{table}

\noindent \textbf{Effectiveness of Cross-Layer Knowledge Transfer.}
We further analyze the effectiveness of Cross-Layer Knowledge Transfer in Table~\ref{tab:ablation_clkt}. 
By comparing the two partitioning methods \method{}~(\textbf{w/o.} CLKT, Row~1$\sim$4) and \method{}~(\textbf{w.} CLKT, Row~5$\sim$8), we found that the performance of \method{}~(\textbf{w.} CLKT) is better than \method{}~(\textbf{w/o.} CLKT). \method{}~(\textbf{w.} CLKT) achieves better performance in most cases~(such as Row~2 vs. Row~5, Row~3 vs. Row~6), indicating that CLKT restricts multi-layer features to one same space to ensure partitioning consistency is necessary.
\begin{table}[!h]
\vspace{-0.12cm}
\setlength{\tabcolsep}{2pt}
  \caption{
  Effectiveness of Dynamic Assignment.
  } 
  \vspace{-0.3cm}
  \label{tab:ablation_prototype}
  \centering
 \renewcommand{\arraystretch}{0.96}
  \resizebox{0.475\textwidth}{!}{

    \begin{tabular}{c|c|c|c|c|c|c}
    \toprule[2pt]
\multirow{2}{*}{\textbf{\makecell{Prototype\\Models}}} & \multirow{2}{*}{\textbf{\makecell{Group\\Numbers}}} & \multicolumn{5}{c}{\texttt{Beauty Dataset}} \\

\cline{3-7}
    \multirow{2}{*} & \multirow{2}{*} & \textbf{AUC} $\uparrow$ & \textbf{NDCG@10}  $\uparrow$ & \textbf{HR@10} $\uparrow$ & \textbf{NDCG@20} $\uparrow$ & \textbf{HR@20} $\uparrow$ \\
    \midrule \midrule
\multirow{4}{*}{Global} & \cellcolor{blue!2}2 & \cellcolor{blue!2}0.6676 & \cellcolor{blue!2}0.2487 & \cellcolor{blue!2}0.4133 & \cellcolor{blue!2}0.2710 & \cellcolor{blue!2}0.5040 \\ \cline{2-7}
\multirow{4}{*}{} & \cellcolor{blue!2}3 & \cellcolor{blue!2}0.6683 & \cellcolor{blue!2}0.2508 & \cellcolor{blue!2}0.4140 & \cellcolor{blue!2}0.2737 & \cellcolor{blue!2}0.5047 \\ \cline{2-7}
\multirow{4}{*}{} & \cellcolor{blue!2}5 & \cellcolor{blue!2}0.6676 & \cellcolor{blue!2}0.2514 & \cellcolor{blue!2}0.4190 & \cellcolor{blue!2}0.2755 & \cellcolor{blue!2}0.5067 \\ \cline{2-7}
\multirow{4}{*}{} & \cellcolor{blue!2}10 & \cellcolor{blue!2}0.6654 & \cellcolor{blue!2}0.2525 & \cellcolor{blue!2}0.4127 & \cellcolor{blue!2}0.2763 & \cellcolor{blue!2}0.5094 \\ \cline{1-7}
\multirow{4}{*}{Group} & \cellcolor{blue!5}2 & \cellcolor{blue!5}0.6685 & \cellcolor{blue!5}0.2499 & \cellcolor{blue!5}0.4153 & \cellcolor{blue!5}0.2729 & \cellcolor{blue!5}0.5080 \\ \cline{2-7}
\multirow{4}{*}{} & \cellcolor{blue!5}3 & \cellcolor{blue!5}0.6690 & \cellcolor{blue!5}0.2511 & \cellcolor{blue!5}0.4160 & \cellcolor{blue!5}0.2749 & \cellcolor{blue!5}0.5060 \\ \cline{2-7}
\multirow{4}{*}{} & \cellcolor{blue!5}5 & \cellcolor{blue!5}0.6688 & \cellcolor{blue!5}0.2520 & \cellcolor{blue!5}0.4207 & \cellcolor{blue!5}0.2736 & \cellcolor{blue!5}0.5070 \\ \cline{2-7}
\multirow{4}{*}{} & \cellcolor{blue!5}10 & \cellcolor{blue!5}0.6677 & \cellcolor{blue!5}0.2527 & \cellcolor{blue!5}0.4173 & \cellcolor{blue!5}0.2764 & \cellcolor{blue!5}0.5124 \\ 
\bottomrule[2pt]
\end{tabular}
}
\vspace{-0.12cm}
\end{table}

\noindent \textbf{Effectiveness of Dynamic Assignment.} 
Table~\ref{tab:ablation_prototype} compares the performance of \method{} with Global \moduleA{} and with Group \moduleA{}s. In most cases, \method{} with Group \moduleA{}s has better performance~(such as Row~3 vs. Row~7, Row~4 vs. Row~8), which shows the effectiveness of dynamic assignment.

\subsection{In-Depth Analysis}
\noindent \textbf{Impact of \modelGradient{}'s threshold.} In order to analyze the impact of the threshold of \modelGradient{} $\mathcal{T}$, we fixed the number of $\mathcal{N}_\mathcal{M}=10$ in Figure~\ref{fig:grad_thres} and adjusted $\mathcal{T}\in\{0.1, 0.5, 1.0, 5.0\}$. From Figure~\ref{fig:grad_thres}, we have the following findings:
The performance is worst when $\mathcal{T}=0.1$, and then the performance is also improved with the increase of $\mathcal{T}$, and it reaches the optimum when $\mathcal{T}=1.0$, but the performance drops when $\mathcal{T}=5.0$, which verifies our hypothesis. When $\mathcal{T}=0.1$, because $\mathcal{T}$ is too small, the Group \moduleB{} cannot personalize well for all samples belonging to this group. When $\mathcal{T}$ increases, the Group \moduleB{} can personalize well for more samples of this group. But when $\mathcal{T}$ is too large, the personalization of the samples in this group becomes unstable due to the large search space of the \modelGradient{}.

\section{Conclusion}
\label{sec:conclusion}

This paper presents \method{}, a novel approach to address real-time distribution shifts on devices without retraining. Using a prototype-based, backpropagation-free parameter editing framework, \method{} adapts models to current data distributions through a cloud-based neural adapter. This approach ensures consistent and context-aware parameter adjustments and efficient prototype model evolution. Extensive experiments on two modalities confirm \method{}'s effectiveness and generality, enhancing on-device model practicality in dynamic environments.

\section*{ACKNOWLEDGEMENTS}
This work was supported by the National Science and Technology Major Project (2022ZD0119100), National Natural Science Foundation of China (62402429, 62441236, U24A20326, 62376243, 62441605, 62037001), the Key Research and Development Program of Zhejiang Province (2025C01026, 2024C03270), the Starry Night Science Fund at Shanghai Institute for Advanced Study (Zhejiang University), Ningbo Yongjiang Talent Introduction Programme (2023A-397-G), Young Elite Scientists Sponsorship Program by CAST (2024QNRC001), This work was also supported by Ant group.
The author gratefully acknowledges the support of Zhejiang University Education Foundation Qizhen Scholar Foundation.
\clearpage
\bibliographystyle{ACM-Reference-Format}
\balance
\bibliography{reference}


\begin{thebibliography}{74}


\ifx \showCODEN    \undefined \def \showCODEN     #1{\unskip}     \fi
\ifx \showISBNx    \undefined \def \showISBNx     #1{\unskip}     \fi
\ifx \showISBNxiii \undefined \def \showISBNxiii  #1{\unskip}     \fi
\ifx \showISSN     \undefined \def \showISSN      #1{\unskip}     \fi
\ifx \showLCCN     \undefined \def \showLCCN      #1{\unskip}     \fi
\ifx \shownote     \undefined \def \shownote      #1{#1}          \fi
\ifx \showarticletitle \undefined \def \showarticletitle #1{#1}   \fi
\ifx \showURL      \undefined \def \showURL       {\relax}        \fi
\providecommand\bibfield[2]{#2}
\providecommand\bibinfo[2]{#2}
\providecommand\natexlab[1]{#1}
\providecommand\showeprint[2][]{arXiv:#2}

\bibitem[Alaluf et~al\mbox{.}(2022)]%
        {ref:hypernetwork_hyperstyle}
\bibfield{author}{\bibinfo{person}{Yuval Alaluf}, \bibinfo{person}{Omer Tov}, \bibinfo{person}{Ron Mokady}, \bibinfo{person}{Rinon Gal}, {and} \bibinfo{person}{Amit Bermano}.} \bibinfo{year}{2022}\natexlab{}.
\newblock \showarticletitle{Hyperstyle: Stylegan inversion with hypernetworks for real image editing}. In \bibinfo{booktitle}{\emph{Proceedings of the IEEE/CVF Conference on Computer Vision and Pattern Recognition}}. \bibinfo{pages}{18511--18521}.
\newblock


\bibitem[Chen et~al\mbox{.}(2019)]%
        {ref:multi_modal_personalized}
\bibfield{author}{\bibinfo{person}{Xu Chen}, \bibinfo{person}{Hanxiong Chen}, \bibinfo{person}{Hongteng Xu}, \bibinfo{person}{Yongfeng Zhang}, \bibinfo{person}{Yixin Cao}, \bibinfo{person}{Zheng Qin}, {and} \bibinfo{person}{Hongyuan Zha}.} \bibinfo{year}{2019}\natexlab{}.
\newblock \showarticletitle{Personalized fashion recommendation with visual explanations based on multimodal attention network: Towards visually explainable recommendation}. In \bibinfo{booktitle}{\emph{Proceedings of the 42nd International ACM SIGIR Conference on Research and Development in Information Retrieval}}. \bibinfo{pages}{765--774}.
\newblock


\bibitem[Chen et~al\mbox{.}(2020)]%
        {ref:dfn_dynamic_conv}
\bibfield{author}{\bibinfo{person}{Yinpeng Chen}, \bibinfo{person}{Xiyang Dai}, \bibinfo{person}{Mengchen Liu}, \bibinfo{person}{Dongdong Chen}, \bibinfo{person}{Lu Yuan}, {and} \bibinfo{person}{Zicheng Liu}.} \bibinfo{year}{2020}\natexlab{}.
\newblock \showarticletitle{Dynamic convolution: Attention over convolution kernels}. In \bibinfo{booktitle}{\emph{Proceedings of the IEEE/CVF Conference on Computer Vision and Pattern Recognition}}. \bibinfo{pages}{11030--11039}.
\newblock


\bibitem[Ding et~al\mbox{.}(2023)]%
        {ref:devicefinetune2}
\bibfield{author}{\bibinfo{person}{Yucheng Ding}, \bibinfo{person}{Chaoyue Niu}, \bibinfo{person}{Fan Wu}, \bibinfo{person}{Shaojie Tang}, \bibinfo{person}{Chengfei Lyu}, {and} \bibinfo{person}{Guihai Chen}.} \bibinfo{year}{2023}\natexlab{}.
\newblock \showarticletitle{DC-CCL: Device-Cloud Collaborative Controlled Learning for Large Vision Models}.
\newblock \bibinfo{journal}{\emph{arXiv preprint arXiv:2303.10361}} (\bibinfo{year}{2023}).
\newblock


\bibitem[Dinh et~al\mbox{.}(2022)]%
        {ref:hypernetwork_hyperinverter}
\bibfield{author}{\bibinfo{person}{Tan~M Dinh}, \bibinfo{person}{Anh~Tuan Tran}, \bibinfo{person}{Rang Nguyen}, {and} \bibinfo{person}{Binh-Son Hua}.} \bibinfo{year}{2022}\natexlab{}.
\newblock \showarticletitle{Hyperinverter: Improving stylegan inversion via hypernetwork}. In \bibinfo{booktitle}{\emph{Proceedings of the IEEE/CVF Conference on Computer Vision and Pattern Recognition}}. \bibinfo{pages}{11389--11398}.
\newblock


\bibitem[Fu et~al\mbox{.}(2025)]%
        {fu2025forward}
\bibfield{author}{\bibinfo{person}{Kairui Fu}, \bibinfo{person}{Zheqi Lv}, \bibinfo{person}{Shengyu Zhang}, \bibinfo{person}{Fan Wu}, {and} \bibinfo{person}{Kun Kuang}.} \bibinfo{year}{2025}\natexlab{}.
\newblock \showarticletitle{Forward Once for All: Structural Parameterized Adaptation for Efficient Cloud-coordinated On-device Recommendation}. In \bibinfo{booktitle}{\emph{Proceedings of the 31st ACM SIGKDD Conference on Knowledge Discovery and Data Mining V. 1}}. \bibinfo{pages}{318--329}.
\newblock


\bibitem[Fu et~al\mbox{.}(2024)]%
        {fu2024diet}
\bibfield{author}{\bibinfo{person}{Kairui Fu}, \bibinfo{person}{Shengyu Zhang}, \bibinfo{person}{Zheqi Lv}, \bibinfo{person}{Jingyuan Chen}, {and} \bibinfo{person}{Jiwei Li}.} \bibinfo{year}{2024}\natexlab{}.
\newblock \showarticletitle{DIET: Customized Slimming for Incompatible Networks in Sequential Recommendation}. In \bibinfo{booktitle}{\emph{Proceedings of the 30th ACM SIGKDD Conference on Knowledge Discovery and Data Mining}}.
\newblock


\bibitem[Gan et~al\mbox{.}(2023a)]%
        {ref:device_cloud_cv}
\bibfield{author}{\bibinfo{person}{Yulu Gan}, \bibinfo{person}{Mingjie Pan}, \bibinfo{person}{Rongyu Zhang}, \bibinfo{person}{Zijian Ling}, \bibinfo{person}{Lingran Zhao}, \bibinfo{person}{Jiaming Liu}, {and} \bibinfo{person}{Shanghang Zhang}.} \bibinfo{year}{2023}\natexlab{a}.
\newblock \showarticletitle{Cloud-device collaborative adaptation to continual changing environments in the real-world}. In \bibinfo{booktitle}{\emph{Proceedings of the IEEE/CVF Conference on Computer Vision and Pattern Recognition}}. \bibinfo{pages}{12157--12166}.
\newblock


\bibitem[Gan et~al\mbox{.}(2023b)]%
        {ref:device-cloud_adaptation}
\bibfield{author}{\bibinfo{person}{Yulu Gan}, \bibinfo{person}{Mingjie Pan}, \bibinfo{person}{Rongyu Zhang}, \bibinfo{person}{Zijian Ling}, \bibinfo{person}{Lingran Zhao}, \bibinfo{person}{Jiaming Liu}, {and} \bibinfo{person}{Shanghang Zhang}.} \bibinfo{year}{2023}\natexlab{b}.
\newblock \showarticletitle{Cloud-device collaborative adaptation to continual changing environments in the real-world}. In \bibinfo{booktitle}{\emph{Proceedings of the IEEE/CVF Conference on Computer Vision and Pattern Recognition}}. \bibinfo{pages}{12157--12166}.
\newblock


\bibitem[Guo et~al\mbox{.}(2017)]%
        {ref:deepfm}
\bibfield{author}{\bibinfo{person}{Huifeng Guo}, \bibinfo{person}{Ruiming Tang}, \bibinfo{person}{Yunming Ye}, \bibinfo{person}{Zhenguo Li}, {and} \bibinfo{person}{Xiuqiang He}.} \bibinfo{year}{2017}\natexlab{}.
\newblock \showarticletitle{DeepFM: a factorization-machine based neural network for CTR prediction}.
\newblock \bibinfo{journal}{\emph{International Joint Conference on Artificial Intelligence}} (\bibinfo{year}{2017}).
\newblock


\bibitem[Han et~al\mbox{.}(2020)]%
        {ref:ghostnet}
\bibfield{author}{\bibinfo{person}{Kai Han}, \bibinfo{person}{Yunhe Wang}, \bibinfo{person}{Qi Tian}, \bibinfo{person}{Jianyuan Guo}, \bibinfo{person}{Chunjing Xu}, {and} \bibinfo{person}{Chang Xu}.} \bibinfo{year}{2020}\natexlab{}.
\newblock \showarticletitle{Ghostnet: More features from cheap operations}. In \bibinfo{booktitle}{\emph{Proceedings of the IEEE/CVF Conference on Computer Vision and Pattern Recognition}}. \bibinfo{pages}{1580--1589}.
\newblock


\bibitem[He et~al\mbox{.}(2019)]%
        {ref:dfn_semantic_segmentation}
\bibfield{author}{\bibinfo{person}{Junjun He}, \bibinfo{person}{Zhongying Deng}, {and} \bibinfo{person}{Yu Qiao}.} \bibinfo{year}{2019}\natexlab{}.
\newblock \showarticletitle{Dynamic multi-scale filters for semantic segmentation}. In \bibinfo{booktitle}{\emph{Proceedings of the IEEE/CVF International Conference on Computer Vision}}. \bibinfo{pages}{3562--3572}.
\newblock


\bibitem[He et~al\mbox{.}(2016a)]%
        {ref:resnet}
\bibfield{author}{\bibinfo{person}{Kaiming He}, \bibinfo{person}{Xiangyu Zhang}, \bibinfo{person}{Shaoqing Ren}, {and} \bibinfo{person}{Jian Sun}.} \bibinfo{year}{2016}\natexlab{a}.
\newblock \showarticletitle{Deep residual learning for image recognition}. In \bibinfo{booktitle}{\emph{Proceedings of the IEEE conference on computer vision and pattern recognition}}. \bibinfo{pages}{770--778}.
\newblock


\bibitem[He et~al\mbox{.}(2016b)]%
        {ref:he2016resnet}
\bibfield{author}{\bibinfo{person}{Kaiming He}, \bibinfo{person}{Xiangyu Zhang}, \bibinfo{person}{Shaoqing Ren}, {and} \bibinfo{person}{Jian Sun}.} \bibinfo{year}{2016}\natexlab{b}.
\newblock \showarticletitle{Deep residual learning for image recognition}. In \bibinfo{booktitle}{\emph{Proceedings of the IEEE conference on computer vision and pattern recognition}}. \bibinfo{pages}{770--778}.
\newblock


\bibitem[He et~al\mbox{.}(2020)]%
        {ref:he2020lightgcn}
\bibfield{author}{\bibinfo{person}{Xiangnan He}, \bibinfo{person}{Kuan Deng}, \bibinfo{person}{Xiang Wang}, \bibinfo{person}{Yan Li}, \bibinfo{person}{Yongdong Zhang}, {and} \bibinfo{person}{Meng Wang}.} \bibinfo{year}{2020}\natexlab{}.
\newblock \showarticletitle{Lightgcn: Simplifying and powering graph convolution network for recommendation}. In \bibinfo{booktitle}{\emph{Proceedings of the 43rd International ACM SIGIR conference on research and development in Information Retrieval}}. \bibinfo{pages}{639--648}.
\newblock


\bibitem[Hidasi et~al\mbox{.}(2016)]%
        {ref:gru4rec}
\bibfield{author}{\bibinfo{person}{Bal{\'a}zs Hidasi}, \bibinfo{person}{Alexandros Karatzoglou}, \bibinfo{person}{Linas Baltrunas}, {and} \bibinfo{person}{Domonkos Tikk}.} \bibinfo{year}{2016}\natexlab{}.
\newblock \showarticletitle{Session-based recommendations with recurrent neural networks}.
\newblock \bibinfo{journal}{\emph{International Conference on Learning Representations 2016}} (\bibinfo{year}{2016}).
\newblock


\bibitem[Howard et~al\mbox{.}(2019)]%
        {ref:mobilenetv3}
\bibfield{author}{\bibinfo{person}{Andrew Howard}, \bibinfo{person}{Mark Sandler}, \bibinfo{person}{Grace Chu}, \bibinfo{person}{Liang-Chieh Chen}, \bibinfo{person}{Bo Chen}, \bibinfo{person}{Mingxing Tan}, \bibinfo{person}{Weijun Wang}, \bibinfo{person}{Yukun Zhu}, \bibinfo{person}{Ruoming Pang}, \bibinfo{person}{Vijay Vasudevan}, {et~al\mbox{.}}} \bibinfo{year}{2019}\natexlab{}.
\newblock \showarticletitle{Searching for mobilenetv3}. In \bibinfo{booktitle}{\emph{Proceedings of the IEEE/CVF International Conference on Computer Vision}}. \bibinfo{pages}{1314--1324}.
\newblock


\bibitem[Howard et~al\mbox{.}(2017)]%
        {ref:mobilenet}
\bibfield{author}{\bibinfo{person}{Andrew~G. Howard}, \bibinfo{person}{Menglong Zhu}, \bibinfo{person}{Bo Chen}, \bibinfo{person}{Dmitry Kalenichenko}, \bibinfo{person}{Weijun Wang}, \bibinfo{person}{Tobias Weyand}, \bibinfo{person}{Marco Andreetto}, {and} \bibinfo{person}{Hartwig Adam}.} \bibinfo{year}{2017}\natexlab{}.
\newblock \showarticletitle{MobileNets: Efficient Convolutional Neural Networks for Mobile Vision Applications}.
\newblock \bibinfo{journal}{\emph{CoRR}}  \bibinfo{volume}{abs/1704.04861} (\bibinfo{year}{2017}).
\newblock
\showeprint[arXiv]{1704.04861}
\urldef\tempurl%
\url{http://arxiv.org/abs/1704.04861}
\showURL{%
\tempurl}


\bibitem[Howard and Ruder(2018)]%
        {ref:finetune}
\bibfield{author}{\bibinfo{person}{Jeremy Howard} {and} \bibinfo{person}{Sebastian Ruder}.} \bibinfo{year}{2018}\natexlab{}.
\newblock \showarticletitle{Universal language model fine-tuning for text classification}. In \bibinfo{booktitle}{\emph{ACL 2018-56th Annual Meeting of the Association for Computational Linguistics, Proceedings of the Conference (Long Papers)}}, Vol.~\bibinfo{volume}{1}. Association for Computational Linguistics, \bibinfo{pages}{328--339}.
\newblock


\bibitem[Iandola et~al\mbox{.}(2016)]%
        {ref:squeezenet}
\bibfield{author}{\bibinfo{person}{Forrest~N Iandola}, \bibinfo{person}{Song Han}, \bibinfo{person}{Matthew~W Moskewicz}, \bibinfo{person}{Khalid Ashraf}, \bibinfo{person}{William~J Dally}, {and} \bibinfo{person}{Kurt Keutzer}.} \bibinfo{year}{2016}\natexlab{}.
\newblock \showarticletitle{SqueezeNet: AlexNet-level accuracy with 50x fewer parameters and< 0.5 MB model size}.
\newblock \bibinfo{journal}{\emph{arXiv preprint arXiv:1602.07360}} (\bibinfo{year}{2016}).
\newblock


\bibitem[Ji et~al\mbox{.}(2025)]%
        {ji2025backpropagation}
\bibfield{author}{\bibinfo{person}{Wei Ji}, \bibinfo{person}{Li Li}, \bibinfo{person}{Zheqi Lv}, \bibinfo{person}{Wenqiao Zhang}, \bibinfo{person}{Mengze Li}, \bibinfo{person}{Zhen Wan}, \bibinfo{person}{Wenqiang Lei}, {and} \bibinfo{person}{Roger Zimmermann}.} \bibinfo{year}{2025}\natexlab{}.
\newblock \showarticletitle{Backpropagation-free multi-modal on-device model adaptation via cloud-device collaboration}.
\newblock \bibinfo{journal}{\emph{ACM Transactions on Multimedia Computing, Communications and Applications}} \bibinfo{volume}{21}, \bibinfo{number}{2} (\bibinfo{year}{2025}), \bibinfo{pages}{1--17}.
\newblock


\bibitem[Jia et~al\mbox{.}(2016)]%
        {ref:dfn}
\bibfield{author}{\bibinfo{person}{Xu Jia}, \bibinfo{person}{Bert De~Brabandere}, \bibinfo{person}{Tinne Tuytelaars}, {and} \bibinfo{person}{Luc~V Gool}.} \bibinfo{year}{2016}\natexlab{}.
\newblock \showarticletitle{Dynamic filter networks}.
\newblock \bibinfo{journal}{\emph{Advances in neural information processing systems}}  \bibinfo{volume}{29} (\bibinfo{year}{2016}).
\newblock


\bibitem[Jiang et~al\mbox{.}(2023)]%
        {ref:devicefinetune3}
\bibfield{author}{\bibinfo{person}{Penghao Jiang}, \bibinfo{person}{Ke Xin}, \bibinfo{person}{Chunxi Li}, {and} \bibinfo{person}{Yinsi Zhou}.} \bibinfo{year}{2023}\natexlab{}.
\newblock \showarticletitle{High-efficiency Device-Cloud Collaborative Transformer Model}. In \bibinfo{booktitle}{\emph{Proceedings of the IEEE/CVF Conference on Computer Vision and Pattern Recognition}}. \bibinfo{pages}{2203--2209}.
\newblock


\bibitem[Kang and McAuley(2018)]%
        {ref:sasrec}
\bibfield{author}{\bibinfo{person}{Wang-Cheng Kang} {and} \bibinfo{person}{Julian McAuley}.} \bibinfo{year}{2018}\natexlab{}.
\newblock \showarticletitle{Self-attentive sequential recommendation}. In \bibinfo{booktitle}{\emph{2018 IEEE International Conference on Data Mining (ICDM)}}. IEEE, \bibinfo{pages}{197--206}.
\newblock


\bibitem[Khani et~al\mbox{.}(2023)]%
        {khani2023recl}
\bibfield{author}{\bibinfo{person}{Mehrdad Khani}, \bibinfo{person}{Ganesh Ananthanarayanan}, \bibinfo{person}{Kevin Hsieh}, \bibinfo{person}{Junchen Jiang}, \bibinfo{person}{Ravi Netravali}, \bibinfo{person}{Yuanchao Shu}, \bibinfo{person}{Mohammad Alizadeh}, {and} \bibinfo{person}{Victor Bahl}.} \bibinfo{year}{2023}\natexlab{}.
\newblock \showarticletitle{$\{$RECL$\}$: Responsive $\{$Resource-Efficient$\}$ continuous learning for video analytics}. In \bibinfo{booktitle}{\emph{20th USENIX Symposium on Networked Systems Design and Implementation (NSDI 23)}}. \bibinfo{pages}{917--932}.
\newblock


\bibitem[Li et~al\mbox{.}(2023)]%
        {li2023your_mm}
\bibfield{author}{\bibinfo{person}{Haoxuan Li}, \bibinfo{person}{Yi Bin}, \bibinfo{person}{Junrong Liao}, \bibinfo{person}{Yang Yang}, {and} \bibinfo{person}{Heng~Tao Shen}.} \bibinfo{year}{2023}\natexlab{}.
\newblock \showarticletitle{Your negative may not be true negative: Boosting image-text matching with false negative elimination}. In \bibinfo{booktitle}{\emph{Proceedings of the 31st ACM international conference on multimedia}}. \bibinfo{pages}{924--934}.
\newblock


\bibitem[Li et~al\mbox{.}(2025)]%
        {Li_2025_CVPR_ood}
\bibfield{author}{\bibinfo{person}{Shawn Li}, \bibinfo{person}{Huixian Gong}, \bibinfo{person}{Hao Dong}, \bibinfo{person}{Tiankai Yang}, \bibinfo{person}{Zhengzhong Tu}, {and} \bibinfo{person}{Yue Zhao}.} \bibinfo{year}{2025}\natexlab{}.
\newblock \showarticletitle{DPU: Dynamic Prototype Updating for Multimodal Out-of-Distribution Detection}. In \bibinfo{booktitle}{\emph{Proceedings of the Computer Vision and Pattern Recognition Conference (CVPR)}}. \bibinfo{pages}{10193--10202}.
\newblock


\bibitem[Liang et~al\mbox{.}(2022)]%
        {ref:device-cloud_yolo}
\bibfield{author}{\bibinfo{person}{Siyuan Liang}, \bibinfo{person}{Hao Wu}, \bibinfo{person}{Li Zhen}, \bibinfo{person}{Qiaozhi Hua}, \bibinfo{person}{Sahil Garg}, \bibinfo{person}{Georges Kaddoum}, \bibinfo{person}{Mohammad~Mehedi Hassan}, {and} \bibinfo{person}{Keping Yu}.} \bibinfo{year}{2022}\natexlab{}.
\newblock \showarticletitle{Edge YOLO: Real-time intelligent object detection system based on edge-cloud cooperation in autonomous vehicles}.
\newblock \bibinfo{journal}{\emph{IEEE Transactions on Intelligent Transportation Systems}} \bibinfo{volume}{23}, \bibinfo{number}{12} (\bibinfo{year}{2022}), \bibinfo{pages}{25345--25360}.
\newblock


\bibitem[Lin et~al\mbox{.}(2022)]%
        {ref:devicefinetune}
\bibfield{author}{\bibinfo{person}{Ji Lin}, \bibinfo{person}{Ligeng Zhu}, \bibinfo{person}{Wei-Ming Chen}, \bibinfo{person}{Wei-Chen Wang}, \bibinfo{person}{Chuang Gan}, {and} \bibinfo{person}{Song Han}.} \bibinfo{year}{2022}\natexlab{}.
\newblock \showarticletitle{On-device training under 256kb memory}.
\newblock \bibinfo{journal}{\emph{Advances in Neural Information Processing Systems}}  \bibinfo{volume}{35} (\bibinfo{year}{2022}), \bibinfo{pages}{22941--22954}.
\newblock


\bibitem[Liu et~al\mbox{.}(2023)]%
        {liu2023category_prototype}
\bibfield{author}{\bibinfo{person}{Kai Liu}, \bibinfo{person}{Zhihang Fu}, \bibinfo{person}{Chao Chen}, \bibinfo{person}{Sheng Jin}, \bibinfo{person}{Ze Chen}, \bibinfo{person}{Mingyuan Tao}, \bibinfo{person}{Rongxin Jiang}, {and} \bibinfo{person}{Jieping Ye}.} \bibinfo{year}{2023}\natexlab{}.
\newblock \showarticletitle{Category-extensible out-of-distribution detection via hierarchical context descriptions}.
\newblock \bibinfo{journal}{\emph{Advances in Neural Information Processing Systems}}  \bibinfo{volume}{36} (\bibinfo{year}{2023}), \bibinfo{pages}{33241--33261}.
\newblock


\bibitem[Liu et~al\mbox{.}(2024)]%
        {liu2024rethinking_ood}
\bibfield{author}{\bibinfo{person}{Kai Liu}, \bibinfo{person}{Zhihang Fu}, \bibinfo{person}{Sheng Jin}, \bibinfo{person}{Chao Chen}, \bibinfo{person}{Ze Chen}, \bibinfo{person}{Rongxin Jiang}, \bibinfo{person}{Fan Zhou}, \bibinfo{person}{Yaowu Chen}, {and} \bibinfo{person}{Jieping Ye}.} \bibinfo{year}{2024}\natexlab{}.
\newblock \showarticletitle{Rethinking Out-of-Distribution Detection on Imbalanced Data Distribution}. In \bibinfo{booktitle}{\emph{The Thirty-eighth Annual Conference on Neural Information Processing Systems}}.
\newblock


\bibitem[Liu et~al\mbox{.}(2019)]%
        {ref:liu2019roberta}
\bibfield{author}{\bibinfo{person}{Yinhan Liu}, \bibinfo{person}{Myle Ott}, \bibinfo{person}{Naman Goyal}, \bibinfo{person}{Jingfei Du}, \bibinfo{person}{Mandar Joshi}, \bibinfo{person}{Danqi Chen}, \bibinfo{person}{Omer Levy}, \bibinfo{person}{Mike Lewis}, \bibinfo{person}{Luke Zettlemoyer}, {and} \bibinfo{person}{Veselin Stoyanov}.} \bibinfo{year}{2019}\natexlab{}.
\newblock \showarticletitle{Roberta: A robustly optimized bert pretraining approach}.
\newblock \bibinfo{journal}{\emph{arXiv preprint arXiv:1907.11692}} (\bibinfo{year}{2019}).
\newblock


\bibitem[Lv et~al\mbox{.}(2022)]%
        {ref:device_cloud_walle}
\bibfield{author}{\bibinfo{person}{Chengfei Lv}, \bibinfo{person}{Chaoyue Niu}, \bibinfo{person}{Renjie Gu}, \bibinfo{person}{Xiaotang Jiang}, \bibinfo{person}{Zhaode Wang}, \bibinfo{person}{Bin Liu}, \bibinfo{person}{Ziqi Wu}, \bibinfo{person}{Qiulin Yao}, \bibinfo{person}{Congyu Huang}, \bibinfo{person}{Panos Huang}, {et~al\mbox{.}}} \bibinfo{year}{2022}\natexlab{}.
\newblock \showarticletitle{Walle: An End-to-End, General-Purpose, and Large-Scale Production System for Device-Cloud Collaborative Machine Learning}.
\newblock \bibinfo{journal}{\emph{arXiv preprint arXiv:2205.14833}} (\bibinfo{year}{2022}).
\newblock


\bibitem[Lv et~al\mbox{.}(2024a)]%
        {lv2024semantic}
\bibfield{author}{\bibinfo{person}{Zheqi Lv}, \bibinfo{person}{Shaoxuan He}, \bibinfo{person}{Tianyu Zhan}, \bibinfo{person}{Shengyu Zhang}, \bibinfo{person}{Wenqiao Zhang}, \bibinfo{person}{Jingyuan Chen}, \bibinfo{person}{Zhou Zhao}, {and} \bibinfo{person}{Fei Wu}.} \bibinfo{year}{2024}\natexlab{a}.
\newblock \showarticletitle{Semantic Codebook Learning for Dynamic Recommendation Models}. In \bibinfo{booktitle}{\emph{Proceedings of the 32nd ACM International Conference on Multimedia}}.
\newblock


\bibitem[Lv et~al\mbox{.}(2025)]%
        {lv2025collaboration}
\bibfield{author}{\bibinfo{person}{Zheqi Lv}, \bibinfo{person}{Tianyu Zhan}, \bibinfo{person}{Wenjie Wang}, \bibinfo{person}{Xinyu Lin}, \bibinfo{person}{Shengyu Zhang}, \bibinfo{person}{Wenqiao Zhang}, \bibinfo{person}{Jiwei Li}, \bibinfo{person}{Kun Kuang}, {and} \bibinfo{person}{Fei Wu}.} \bibinfo{year}{2025}\natexlab{}.
\newblock \showarticletitle{Collaboration of Large Language Models and Small Recommendation Models for Device-Cloud Recommendation}. In \bibinfo{booktitle}{\emph{Proceedings of the 31st {ACM} {SIGKDD} Conference on Knowledge Discovery and Data Mining, V.1, {KDD} 2025, Toronto, ON, Canada, August 3-7, 2025}}. \bibinfo{publisher}{{ACM}}, \bibinfo{pages}{962--973}.
\newblock


\bibitem[Lv et~al\mbox{.}(2024b)]%
        {ref:lv2024intelligent}
\bibfield{author}{\bibinfo{person}{Zheqi Lv}, \bibinfo{person}{Wenqiao Zhang}, \bibinfo{person}{Zhengyu Chen}, \bibinfo{person}{Shengyu Zhang}, {and} \bibinfo{person}{Kun Kuang}.} \bibinfo{year}{2024}\natexlab{b}.
\newblock \showarticletitle{Intelligent model update strategy for sequential recommendation}. In \bibinfo{booktitle}{\emph{Proceedings of the ACM on Web Conference 2024}}. \bibinfo{pages}{3117--3128}.
\newblock


\bibitem[Lv et~al\mbox{.}(2023)]%
        {ref:duet}
\bibfield{author}{\bibinfo{person}{Zheqi Lv}, \bibinfo{person}{Wenqiao Zhang}, \bibinfo{person}{Shengyu Zhang}, \bibinfo{person}{Kun Kuang}, \bibinfo{person}{Feng Wang}, \bibinfo{person}{Yongwei Wang}, \bibinfo{person}{Zhengyu Chen}, \bibinfo{person}{Tao Shen}, \bibinfo{person}{Hongxia Yang}, \bibinfo{person}{Beng~Chin Ooi}, {and} \bibinfo{person}{Fei Wu}.} \bibinfo{year}{2023}\natexlab{}.
\newblock \showarticletitle{DUET: A Tuning-Free Device-Cloud Collaborative Parameters Generation Framework for Efficient Device Model Generalization}. In \bibinfo{booktitle}{\emph{Proceedings of the ACM Web Conference 2023}}.
\newblock


\bibitem[Ma et~al\mbox{.}(2018)]%
        {ref:shufflenetv2}
\bibfield{author}{\bibinfo{person}{Ningning Ma}, \bibinfo{person}{Xiangyu Zhang}, \bibinfo{person}{Hai-Tao Zheng}, {and} \bibinfo{person}{Jian Sun}.} \bibinfo{year}{2018}\natexlab{}.
\newblock \showarticletitle{Shufflenet v2: Practical guidelines for efficient cnn architecture design}. In \bibinfo{booktitle}{\emph{Proceedings of the European conference on computer vision (ECCV)}}. \bibinfo{pages}{116--131}.
\newblock


\bibitem[Mazari et~al\mbox{.}(2024)]%
        {ref:ensemble_mazari2024bert}
\bibfield{author}{\bibinfo{person}{Ahmed~Cherif Mazari}, \bibinfo{person}{Nesrine Boudoukhani}, {and} \bibinfo{person}{Abdelhamid Djeffal}.} \bibinfo{year}{2024}\natexlab{}.
\newblock \showarticletitle{BERT-based ensemble learning for multi-aspect hate speech detection}.
\newblock \bibinfo{journal}{\emph{Cluster Computing}} \bibinfo{volume}{27}, \bibinfo{number}{1} (\bibinfo{year}{2024}), \bibinfo{pages}{325--339}.
\newblock


\bibitem[Profentzas et~al\mbox{.}(2022)]%
        {ref:devicefinetune4}
\bibfield{author}{\bibinfo{person}{Christos Profentzas}, \bibinfo{person}{Magnus Almgren}, {and} \bibinfo{person}{Olaf Landsiedel}.} \bibinfo{year}{2022}\natexlab{}.
\newblock \showarticletitle{MiniLearn: On-Device Learning for Low-Power IoT Devices.}. In \bibinfo{booktitle}{\emph{EWSN}}. \bibinfo{pages}{1--11}.
\newblock


\bibitem[Qian et~al\mbox{.}(2022)]%
        {ref:zhang_device_cloud}
\bibfield{author}{\bibinfo{person}{Xufeng Qian}, \bibinfo{person}{Yue Xu}, \bibinfo{person}{Fuyu Lv}, \bibinfo{person}{Shengyu Zhang}, \bibinfo{person}{Ziwen Jiang}, \bibinfo{person}{Qingwen Liu}, \bibinfo{person}{Xiaoyi Zeng}, \bibinfo{person}{Tat{-}Seng Chua}, {and} \bibinfo{person}{Fei Wu}.} \bibinfo{year}{2022}\natexlab{}.
\newblock \showarticletitle{Intelligent Request Strategy Design in Recommender System}. In \bibinfo{booktitle}{\emph{{KDD} '22: The 28th {ACM} {SIGKDD} Conference on Knowledge Discovery and Data Mining}}. \bibinfo{publisher}{{ACM}}, \bibinfo{pages}{3772--3782}.
\newblock


\bibitem[Sandler et~al\mbox{.}(2018)]%
        {ref:mobilenetv2}
\bibfield{author}{\bibinfo{person}{Mark Sandler}, \bibinfo{person}{Andrew Howard}, \bibinfo{person}{Menglong Zhu}, \bibinfo{person}{Andrey Zhmoginov}, {and} \bibinfo{person}{Liang-Chieh Chen}.} \bibinfo{year}{2018}\natexlab{}.
\newblock \showarticletitle{Mobilenetv2: Inverted residuals and linear bottlenecks}. In \bibinfo{booktitle}{\emph{Proceedings of the IEEE conference on computer vision and pattern recognition}}. \bibinfo{pages}{4510--4520}.
\newblock


\bibitem[Shamsian et~al\mbox{.}(2021)]%
        {ref:hypernetwork_federated_learning}
\bibfield{author}{\bibinfo{person}{Aviv Shamsian}, \bibinfo{person}{Aviv Navon}, \bibinfo{person}{Ethan Fetaya}, {and} \bibinfo{person}{Gal Chechik}.} \bibinfo{year}{2021}\natexlab{}.
\newblock \showarticletitle{Personalized federated learning using hypernetworks}. In \bibinfo{booktitle}{\emph{International Conference on Machine Learning}}. PMLR, \bibinfo{pages}{9489--9502}.
\newblock


\bibitem[Simonyan and Zisserman(2015)]%
        {ref:vggnet}
\bibfield{author}{\bibinfo{person}{Karen Simonyan} {and} \bibinfo{person}{Andrew Zisserman}.} \bibinfo{year}{2015}\natexlab{}.
\newblock \showarticletitle{Very Deep Convolutional Networks for Large-Scale Image Recognition}. In \bibinfo{booktitle}{\emph{3rd International Conference on Learning Representations, {ICLR} 2015}}, \bibfield{editor}{\bibinfo{person}{Yoshua Bengio} {and} \bibinfo{person}{Yann LeCun}} (Eds.).
\newblock


\bibitem[Su et~al\mbox{.}(2019)]%
        {ref:dfn_unknown2}
\bibfield{author}{\bibinfo{person}{Hang Su}, \bibinfo{person}{Varun Jampani}, \bibinfo{person}{Deqing Sun}, \bibinfo{person}{Orazio Gallo}, \bibinfo{person}{Erik Learned-Miller}, {and} \bibinfo{person}{Jan Kautz}.} \bibinfo{year}{2019}\natexlab{}.
\newblock \showarticletitle{Pixel-adaptive convolutional neural networks}. In \bibinfo{booktitle}{\emph{Proceedings of the IEEE/CVF Conference on Computer Vision and Pattern Recognition}}. \bibinfo{pages}{11166--11175}.
\newblock


\bibitem[Su et~al\mbox{.}(2023a)]%
        {su2023personalized}
\bibfield{author}{\bibinfo{person}{Jiajie Su}, \bibinfo{person}{Chaochao Chen}, \bibinfo{person}{Zibin Lin}, \bibinfo{person}{Xi Li}, \bibinfo{person}{Weiming Liu}, {and} \bibinfo{person}{Xiaolin Zheng}.} \bibinfo{year}{2023}\natexlab{a}.
\newblock \showarticletitle{Personalized behavior-aware transformer for multi-behavior sequential recommendation}. In \bibinfo{booktitle}{\emph{Proceedings of the 31st ACM international conference on multimedia}}. \bibinfo{pages}{6321--6331}.
\newblock


\bibitem[Su et~al\mbox{.}(2023b)]%
        {su2023enhancing}
\bibfield{author}{\bibinfo{person}{Jiajie Su}, \bibinfo{person}{Chaochao Chen}, \bibinfo{person}{Weiming Liu}, \bibinfo{person}{Fei Wu}, \bibinfo{person}{Xiaolin Zheng}, {and} \bibinfo{person}{Haoming Lyu}.} \bibinfo{year}{2023}\natexlab{b}.
\newblock \showarticletitle{Enhancing hierarchy-aware graph networks with deep dual clustering for session-based recommendation}. In \bibinfo{booktitle}{\emph{Proceedings of the ACM web conference 2023}}. \bibinfo{pages}{165--176}.
\newblock


\bibitem[Su et~al\mbox{.}(2025)]%
        {su2025distilling_mmrec}
\bibfield{author}{\bibinfo{person}{Jiajie Su}, \bibinfo{person}{Qiyong Zhong}, \bibinfo{person}{Yunshan Ma}, \bibinfo{person}{Weiming Liu}, \bibinfo{person}{Chaochao Chen}, \bibinfo{person}{Xiaolin Zheng}, \bibinfo{person}{Jianwei Yin}, {and} \bibinfo{person}{Tat-Seng Chua}.} \bibinfo{year}{2025}\natexlab{}.
\newblock \showarticletitle{Distilling Transitional Pattern to Large Language Models for Multimodal Session-based Recommendation}.
\newblock \bibinfo{journal}{\emph{arXiv preprint arXiv:2504.10538}} (\bibinfo{year}{2025}).
\newblock


\bibitem[Sun et~al\mbox{.}(2019)]%
        {ref:bert4rec}
\bibfield{author}{\bibinfo{person}{Fei Sun}, \bibinfo{person}{Jun Liu}, \bibinfo{person}{Jian Wu}, \bibinfo{person}{Changhua Pei}, \bibinfo{person}{Xiao Lin}, \bibinfo{person}{Wenwu Ou}, {and} \bibinfo{person}{Peng Jiang}.} \bibinfo{year}{2019}\natexlab{}.
\newblock \showarticletitle{BERT4Rec: Sequential recommendation with bidirectional encoder representations from transformer}. In \bibinfo{booktitle}{\emph{Proceedings of the 28th ACM international conference on information and knowledge management}}. \bibinfo{pages}{1441--1450}.
\newblock


\bibitem[Tan and Le(2019)]%
        {ref:efficientnet}
\bibfield{author}{\bibinfo{person}{Mingxing Tan} {and} \bibinfo{person}{Quoc Le}.} \bibinfo{year}{2019}\natexlab{}.
\newblock \showarticletitle{Efficientnet: Rethinking model scaling for convolutional neural networks}. In \bibinfo{booktitle}{\emph{International conference on machine learning}}. PMLR, \bibinfo{pages}{6105--6114}.
\newblock


\bibitem[Tang et~al\mbox{.}(2024)]%
        {tang2024modelgpt}
\bibfield{author}{\bibinfo{person}{Zihao Tang}, \bibinfo{person}{Zheqi Lv}, \bibinfo{person}{Shengyu Zhang}, \bibinfo{person}{Fei Wu}, {and} \bibinfo{person}{Kun Kuang}.} \bibinfo{year}{2024}\natexlab{}.
\newblock \showarticletitle{ModelGPT: Unleashing LLM's Capabilities for Tailored Model Generation}.
\newblock \bibinfo{journal}{\emph{arXiv preprint arXiv:2402.12408}} (\bibinfo{year}{2024}).
\newblock


\bibitem[Tong et~al\mbox{.}(2023)]%
        {HTCL_ood}
\bibfield{author}{\bibinfo{person}{Yunze Tong}, \bibinfo{person}{Junkun Yuan}, \bibinfo{person}{Min Zhang}, \bibinfo{person}{Didi Zhu}, \bibinfo{person}{Keli Zhang}, \bibinfo{person}{Fei Wu}, {and} \bibinfo{person}{Kun Kuang}.} \bibinfo{year}{2023}\natexlab{}.
\newblock \showarticletitle{Quantitatively Measuring and Contrastively Exploring Heterogeneity for Domain Generalization}. In \bibinfo{booktitle}{\emph{Proceedings of the 29th ACM SIGKDD Conference on Knowledge Discovery and Data Mining}}.
\newblock


\bibitem[von Oswald et~al\mbox{.}(2020)]%
        {ref:hypernetwork_continual_learning}
\bibfield{author}{\bibinfo{person}{Johannes von Oswald}, \bibinfo{person}{Christian Henning}, \bibinfo{person}{Jo{\~{a}}o Sacramento}, {and} \bibinfo{person}{Benjamin~F. Grewe}.} \bibinfo{year}{2020}\natexlab{}.
\newblock \showarticletitle{Continual learning with hypernetworks}. In \bibinfo{booktitle}{\emph{8th International Conference on Learning Representations, {ICLR} 2020}}.
\newblock


\bibitem[Wang et~al\mbox{.}(2019)]%
        {ref:dfn_unknown1}
\bibfield{author}{\bibinfo{person}{Jiaqi Wang}, \bibinfo{person}{Kai Chen}, \bibinfo{person}{Rui Xu}, \bibinfo{person}{Ziwei Liu}, \bibinfo{person}{Chen~Change Loy}, {and} \bibinfo{person}{Dahua Lin}.} \bibinfo{year}{2019}\natexlab{}.
\newblock \showarticletitle{Carafe: Content-aware reassembly of features}. In \bibinfo{booktitle}{\emph{Proceedings of the IEEE/CVF International Conference on Computer Vision}}. \bibinfo{pages}{3007--3016}.
\newblock


\bibitem[Wang et~al\mbox{.}(2025)]%
        {wangneural}
\bibfield{author}{\bibinfo{person}{Jiawei Wang}, \bibinfo{person}{Shaofei Lu}, \bibinfo{person}{Da Cao}, \bibinfo{person}{Dongyu Wang}, \bibinfo{person}{Yuquan Le}, \bibinfo{person}{Zhe Quan}, {and} \bibinfo{person}{Tat-Seng Chua}.} \bibinfo{year}{2025}\natexlab{}.
\newblock \showarticletitle{Neural Causal Graph for Interpretable and Intervenable Classification}. In \bibinfo{booktitle}{\emph{The Thirteenth International Conference on Learning Representations}}.
\newblock


\bibitem[Wang et~al\mbox{.}(2023)]%
        {wang2023deconfounded_mm}
\bibfield{author}{\bibinfo{person}{Jiawei Wang}, \bibinfo{person}{Zhanchang Ma}, \bibinfo{person}{Da Cao}, \bibinfo{person}{Yuquan Le}, \bibinfo{person}{Junbin Xiao}, {and} \bibinfo{person}{Tat-Seng Chua}.} \bibinfo{year}{2023}\natexlab{}.
\newblock \showarticletitle{Deconfounded Multimodal Learning for Spatio-temporal Video Grounding}. In \bibinfo{booktitle}{\emph{Proceedings of the 31st ACM International Conference on Multimedia}}. \bibinfo{pages}{7521--7529}.
\newblock


\bibitem[Wang et~al\mbox{.}(2022)]%
        {ref:test_time_adaptation}
\bibfield{author}{\bibinfo{person}{Qin Wang}, \bibinfo{person}{Olga Fink}, \bibinfo{person}{Luc Van~Gool}, {and} \bibinfo{person}{Dengxin Dai}.} \bibinfo{year}{2022}\natexlab{}.
\newblock \showarticletitle{Continual test-time domain adaptation}. In \bibinfo{booktitle}{\emph{Proceedings of the IEEE/CVF Conference on Computer Vision and Pattern Recognition}}. \bibinfo{pages}{7201--7211}.
\newblock


\bibitem[Wang et~al\mbox{.}(2020)]%
        {ref:dfn_instance_segmentation}
\bibfield{author}{\bibinfo{person}{Xinlong Wang}, \bibinfo{person}{Rufeng Zhang}, \bibinfo{person}{Tao Kong}, \bibinfo{person}{Lei Li}, {and} \bibinfo{person}{Chunhua Shen}.} \bibinfo{year}{2020}\natexlab{}.
\newblock \showarticletitle{Solov2: Dynamic and fast instance segmentation}.
\newblock \bibinfo{journal}{\emph{Advances in Neural information processing systems}}  \bibinfo{volume}{33} (\bibinfo{year}{2020}), \bibinfo{pages}{17721--17732}.
\newblock


\bibitem[Wu et~al\mbox{.}(2025)]%
        {wu2025embracing}
\bibfield{author}{\bibinfo{person}{Tao Wu}, \bibinfo{person}{Jingyuan Chen}, \bibinfo{person}{Wang Lin}, \bibinfo{person}{Mengze Li}, \bibinfo{person}{Yumeng Zhu}, \bibinfo{person}{Ang Li}, \bibinfo{person}{Kun Kuang}, {and} \bibinfo{person}{Fei Wu}.} \bibinfo{year}{2025}\natexlab{}.
\newblock \showarticletitle{Embracing Imperfection: Simulating Students with Diverse Cognitive Levels Using LLM-based Agents}.
\newblock \bibinfo{journal}{\emph{arXiv preprint arXiv:2505.19997}} (\bibinfo{year}{2025}).
\newblock


\bibitem[Xian et~al\mbox{.}(2021)]%
        {ref:hypernetwork_meta_learning}
\bibfield{author}{\bibinfo{person}{Zhou Xian}, \bibinfo{person}{Shamit Lal}, \bibinfo{person}{Hsiao{-}Yu Tung}, \bibinfo{person}{Emmanouil~Antonios Platanios}, {and} \bibinfo{person}{Katerina Fragkiadaki}.} \bibinfo{year}{2021}\natexlab{}.
\newblock \showarticletitle{HyperDynamics: Meta-Learning Object and Agent Dynamics with Hypernetworks}. In \bibinfo{booktitle}{\emph{9th International Conference on Learning Representations, {ICLR} 2021}}.
\newblock


\bibitem[Yan et~al\mbox{.}(2022)]%
        {ref:device_cloud2}
\bibfield{author}{\bibinfo{person}{Yikai Yan}, \bibinfo{person}{Chaoyue Niu}, \bibinfo{person}{Renjie Gu}, \bibinfo{person}{Fan Wu}, \bibinfo{person}{Shaojie Tang}, \bibinfo{person}{Lifeng Hua}, \bibinfo{person}{Chengfei Lyu}, {and} \bibinfo{person}{Guihai Chen}.} \bibinfo{year}{2022}\natexlab{}.
\newblock \showarticletitle{On-Device Learning for Model Personalization with Large-Scale Cloud-Coordinated Domain Adaption}. In \bibinfo{booktitle}{\emph{{KDD} '22: The 28th {ACM} {SIGKDD} Conference on Knowledge Discovery and Data Mining, Washington, DC, USA, August 14 - 18, 2022}}. \bibinfo{pages}{2180--2190}.
\newblock


\bibitem[Yang et~al\mbox{.}(2019)]%
        {ref:dfn_condconv}
\bibfield{author}{\bibinfo{person}{Brandon Yang}, \bibinfo{person}{Gabriel Bender}, \bibinfo{person}{Quoc~V Le}, {and} \bibinfo{person}{Jiquan Ngiam}.} \bibinfo{year}{2019}\natexlab{}.
\newblock \showarticletitle{Condconv: Conditionally parameterized convolutions for efficient inference}.
\newblock \bibinfo{journal}{\emph{Advances in Neural Information Processing Systems}}  \bibinfo{volume}{32} (\bibinfo{year}{2019}).
\newblock


\bibitem[Yang et~al\mbox{.}(2024)]%
        {yang2024explain_seqood}
\bibfield{author}{\bibinfo{person}{Linxiao Yang}, \bibinfo{person}{Yunze Tong}, \bibinfo{person}{Xinyue Gu}, {and} \bibinfo{person}{Liang Sun}.} \bibinfo{year}{2024}\natexlab{}.
\newblock \showarticletitle{Explain temporal black-box models via functional decomposition}. In \bibinfo{booktitle}{\emph{Proceedings of the 41st International Conference on Machine Learning}}.
\newblock


\bibitem[Yang et~al\mbox{.}(2022)]%
        {DBLP:conf/kdd/YangHXLYL22}
\bibfield{author}{\bibinfo{person}{Yuhao Yang}, \bibinfo{person}{Chao Huang}, \bibinfo{person}{Lianghao Xia}, \bibinfo{person}{Yuxuan Liang}, \bibinfo{person}{Yanwei Yu}, {and} \bibinfo{person}{Chenliang Li}.} \bibinfo{year}{2022}\natexlab{}.
\newblock \showarticletitle{Multi-Behavior Hypergraph-Enhanced Transformer for Sequential Recommendation}. In \bibinfo{booktitle}{\emph{{KDD} '22: The 28th {ACM} {SIGKDD} Conference on Knowledge Discovery and Data Mining, Washington, DC, USA, August 14 - 18, 2022}}. \bibinfo{publisher}{{ACM}}, \bibinfo{pages}{2263--2274}.
\newblock


\bibitem[Yao et~al\mbox{.}(2022)]%
        {ref:device_cloud}
\bibfield{author}{\bibinfo{person}{Jiangchao Yao}, \bibinfo{person}{Feng Wang}, \bibinfo{person}{Xichen Ding}, \bibinfo{person}{Shaohu Chen}, \bibinfo{person}{Bo Han}, \bibinfo{person}{Jingren Zhou}, {and} \bibinfo{person}{Hongxia Yang}.} \bibinfo{year}{2022}\natexlab{}.
\newblock \showarticletitle{Device-cloud Collaborative Recommendation via Meta Controller}. In \bibinfo{booktitle}{\emph{{KDD} '22: The 28th {ACM} {SIGKDD} Conference on Knowledge Discovery and Data Mining, Washington, DC, USA, August 14 - 18, 2022}}. \bibinfo{pages}{4353--4362}.
\newblock


\bibitem[Yao et~al\mbox{.}(2021)]%
        {ref:dccl}
\bibfield{author}{\bibinfo{person}{Jiangchao Yao}, \bibinfo{person}{Feng Wang}, \bibinfo{person}{Kunyang Jia}, \bibinfo{person}{Bo Han}, \bibinfo{person}{Jingren Zhou}, {and} \bibinfo{person}{Hongxia Yang}.} \bibinfo{year}{2021}\natexlab{}.
\newblock \showarticletitle{Device-cloud collaborative learning for recommendation}. In \bibinfo{booktitle}{\emph{Proceedings of the 27th ACM SIGKDD Conference on Knowledge Discovery \& Data Mining}}. \bibinfo{pages}{3865--3874}.
\newblock


\bibitem[Zhang et~al\mbox{.}(2019)]%
        {ref:hypernetwork_graph}
\bibfield{author}{\bibinfo{person}{Chris Zhang}, \bibinfo{person}{Mengye Ren}, {and} \bibinfo{person}{Raquel Urtasun}.} \bibinfo{year}{2019}\natexlab{}.
\newblock \showarticletitle{Graph HyperNetworks for Neural Architecture Search}. In \bibinfo{booktitle}{\emph{7th International Conference on Learning Representations, {ICLR} 2019}}.
\newblock


\bibitem[Zhang et~al\mbox{.}(2024)]%
        {zhang2024modality}
\bibfield{author}{\bibinfo{person}{Jinghao Zhang}, \bibinfo{person}{Guofan Liu}, \bibinfo{person}{Qiang Liu}, \bibinfo{person}{Shu Wu}, {and} \bibinfo{person}{Liang Wang}.} \bibinfo{year}{2024}\natexlab{}.
\newblock \showarticletitle{Modality-Balanced Learning for Multimedia Recommendation}. In \bibinfo{booktitle}{\emph{Proceedings of the 32nd ACM International Conference on Multimedia}}. \bibinfo{pages}{7551--7560}.
\newblock


\bibitem[Zhang et~al\mbox{.}(2025)]%
        {zhang2025personalized}
\bibfield{author}{\bibinfo{person}{Jinghao Zhang}, \bibinfo{person}{Yuting Liu}, \bibinfo{person}{Wenjie Wang}, \bibinfo{person}{Qiang Liu}, \bibinfo{person}{Shu Wu}, \bibinfo{person}{Liang Wang}, {and} \bibinfo{person}{Tat-Seng Chua}.} \bibinfo{year}{2025}\natexlab{}.
\newblock \showarticletitle{Personalized Text Generation with Contrastive Activation Steering}.
\newblock \bibinfo{journal}{\emph{arXiv preprint arXiv:2503.05213}} (\bibinfo{year}{2025}).
\newblock


\bibitem[Zhang et~al\mbox{.}(2021)]%
        {zhang2021mining_mmrec}
\bibfield{author}{\bibinfo{person}{Jinghao Zhang}, \bibinfo{person}{Yanqiao Zhu}, \bibinfo{person}{Qiang Liu}, \bibinfo{person}{Shu Wu}, \bibinfo{person}{Shuhui Wang}, {and} \bibinfo{person}{Liang Wang}.} \bibinfo{year}{2021}\natexlab{}.
\newblock \showarticletitle{Mining latent structures for multimedia recommendation}. In \bibinfo{booktitle}{\emph{Proceedings of the 29th ACM international conference on multimedia}}. \bibinfo{pages}{3872--3880}.
\newblock


\bibitem[Zhang et~al\mbox{.}(2018)]%
        {ref:shufflenet}
\bibfield{author}{\bibinfo{person}{Xiangyu Zhang}, \bibinfo{person}{Xinyu Zhou}, \bibinfo{person}{Mengxiao Lin}, {and} \bibinfo{person}{Jian Sun}.} \bibinfo{year}{2018}\natexlab{}.
\newblock \showarticletitle{Shufflenet: An extremely efficient convolutional neural network for mobile devices}. In \bibinfo{booktitle}{\emph{Proceedings of the IEEE conference on computer vision and pattern recognition}}. \bibinfo{pages}{6848--6856}.
\newblock


\bibitem[Zhou et~al\mbox{.}(2018)]%
        {ref:din}
\bibfield{author}{\bibinfo{person}{Guorui Zhou}, \bibinfo{person}{Xiaoqiang Zhu}, \bibinfo{person}{Chenru Song}, \bibinfo{person}{Ying Fan}, \bibinfo{person}{Han Zhu}, \bibinfo{person}{Xiao Ma}, \bibinfo{person}{Yanghui Yan}, \bibinfo{person}{Junqi Jin}, \bibinfo{person}{Han Li}, {and} \bibinfo{person}{Kun Gai}.} \bibinfo{year}{2018}\natexlab{}.
\newblock \showarticletitle{Deep interest network for click-through rate prediction}. In \bibinfo{booktitle}{\emph{Proceedings of the 24th ACM SIGKDD International Conference on Knowledge Discovery \& Data Mining}}. \bibinfo{pages}{1059--1068}.
\newblock


\bibitem[Zhou et~al\mbox{.}(2021)]%
        {ref:dfn_lightweight}
\bibfield{author}{\bibinfo{person}{Jingkai Zhou}, \bibinfo{person}{Varun Jampani}, \bibinfo{person}{Zhixiong Pi}, \bibinfo{person}{Qiong Liu}, {and} \bibinfo{person}{Ming-Hsuan Yang}.} \bibinfo{year}{2021}\natexlab{}.
\newblock \showarticletitle{Decoupled dynamic filter networks}. In \bibinfo{booktitle}{\emph{Proceedings of the IEEE/CVF Conference on Computer Vision and Pattern Recognition}}. \bibinfo{pages}{6647--6656}.
\newblock


\bibitem[Zhu et~al\mbox{.}(2025)]%
        {zhu2025graphclip}
\bibfield{author}{\bibinfo{person}{Yun Zhu}, \bibinfo{person}{Haizhou Shi}, \bibinfo{person}{Xiaotang Wang}, \bibinfo{person}{Yongchao Liu}, \bibinfo{person}{Yaoke Wang}, \bibinfo{person}{Boci Peng}, \bibinfo{person}{Chuntao Hong}, {and} \bibinfo{person}{Siliang Tang}.} \bibinfo{year}{2025}\natexlab{}.
\newblock \showarticletitle{Graphclip: Enhancing transferability in graph foundation models for text-attributed graphs}. In \bibinfo{booktitle}{\emph{Proceedings of the ACM on Web Conference 2025}}. \bibinfo{pages}{2183--2197}.
\newblock


\end{thebibliography}
\end{document}